\documentclass[sigconf]{acmart}
\copyrightyear{2026}
\acmYear{2026}
\setcopyright{cc}
\setcctype{by}
\acmConference[WWW '26] {Proceedings of the ACM Web Conference 2026}{April 13--17, 2026}{Dubai, United Arab Emirates.}
\acmBooktitle{Proceedings of the ACM Web Conference 2026 (WWW '26), April 13--17, 2026, Dubai, United Arab Emirates}
\acmISBN{979-8-4007-2307-0/2026/04}
\acmDOI{10.1145/3774904.3792422}
\usepackage{graphicx}
\usepackage{amsmath}
\usepackage{amsthm}
\usepackage{xcolor,colortbl}
\usepackage{siunitx}   % 用于数字对齐
\usepackage{hyperref}  % 先加载 hyperref
\usepackage{multirow}
\usepackage{bm}
\usepackage{subcaption}
\usepackage{enumitem}
\usepackage{mathtools}
\usepackage{algorithm}
\usepackage{algorithmic}

\begin{document}
	%\title{Are Counterfactual Truly Beneficial for Continuous-Time Dynamic Link Prediction?}
	\title{TFWaveFormer: Temporal-Frequency Collaborative Multi-level Wavelet Transformer for Dynamic Link Prediction}
	\settopmatter{authorsperrow=4}
	\author{Hantong Feng}
	\orcid{0000-0002-8593-9281}
	\affiliation{%
		\institution{School of Cyber Science and Engineering, Southeast University}
		\city{Nanjing}
		\country{China}}
	\email{htfeng@seu.edu.cn}
	
	\author{Yonggang Wu}
	\orcid{0009-0004-9468-1333}
	\affiliation{%
		\institution{School of Mathematics, Southeast University}
		\city{Nanjing}
		\country{China}
	}
	\email{wuyg@seu.edu.cn}
	
	\author{Duxin Chen}\authornotemark[1]
	\orcid{0000-0002-3194-2258}
	%\authornote{Corresponding author}
	\affiliation{%
		\institution{School of Mathematics, Southeast University}
		\city{Nanjing}
		\country{China}
	}
	\email{chendx@seu.edu.cn}
	
	\author{Wenwu Yu}
	\orcid{0000-0003-3755-179X}
	%\authornote{Corresponding author}
	\affiliation{%
		\institution{School of Mathematics, Southeast University}
		\city{Nanjing}
		\country{China}
	}
	\email{wwyu@seu.edu.cn}
	
	\authornote{Correspondence to Duxin Chen and Wenwu Yu.}
	%\renewcommand{\shortauthors}{Qijie Bai, Changli Nie, Haiwei Zhang, Dongming Zhao and Xiaojie Yuan}
	
	\begin{abstract}
		%Temporal link prediction in dynamic networks remains a challenging task due to the complex interplay between structural evolution and temporal dependencies. Existing methods often struggle to capture the underlying causal relationships that drive link formation, leading to suboptimal predictive performance. 
		Dynamic link prediction plays a crucial role in diverse applications including social network analysis, communication forecasting, and financial modeling. While recent Transformer-based approaches have demonstrated promising results in temporal graph learning, their performance remains limited when capturing complex multi-scale temporal dynamics. In this paper, we propose TFWaveFormer, a novel Transformer architecture that integrates temporal-frequency analysis with multi-resolution wavelet decomposition to enhance dynamic link prediction. Our framework comprises three key components: (i) a temporal-frequency coordination mechanism that jointly models temporal and spectral representations, (ii) a learnable multi-resolution wavelet decomposition module that adaptively extracts multi-scale temporal patterns through parallel convolutions, replacing traditional iterative wavelet transforms, and (iii) a hybrid Transformer module that effectively fuses local wavelet features with global temporal dependencies. Extensive experiments on benchmark datasets demonstrate that TFWaveFormer achieves state-of-the-art performance, outperforming existing Transformer-based and hybrid models by significant margins across multiple metrics. The superior performance of TFWaveFormer validates the effectiveness of combining temporal-frequency analysis with wavelet decomposition in capturing complex temporal dynamics for dynamic link prediction tasks. %The code is available at \url{https://github.com/SEUFHTong/TFWaveFormer}.
	\end{abstract}

\begin{CCSXML}
    <ccs2012>
    <concept>
    <concept_id>10003752.10003809.10003635.10010038</concept_id>
    <concept_desc>Information systems~Web mining</concept_desc>
    <concept_significance>500</concept_significance>
    </concept>
    </ccs2012>
\end{CCSXML}

\ccsdesc[500]{Theory of computation~Dynamic graph algorithms}

\keywords{Link Prediction; Continuous-Time Dynamic Networks; Wavelet Transforms; Complex Systems}

\maketitle
	
	\section{Introduction}
	Dynamic link prediction, as a core task in temporal graph analysis, derives its importance from the dynamic evolution inherent in real-world networks~\cite{DBLP:journals/jmlr/KazemiGJKSFP20,ekle2024anomaly}. Unlike static graph assumptions, real-world networks, such as social networks~\cite{alvarez2021evolutionary}, biomolecular interaction networks~\cite{DrugInteraction}, and information dissemination networks~\cite{Zhang2020,NBFNet}, exhibit significant temporal variability. Accurately predicting the evolving connections in these networks over time not only holds theoretical value for understanding the dynamic behaviors of complex systems but also provides critical technical support for practical applications, such as recommender systems and epidemic forecasting~\cite{bai2024dynamic,zhang2024llm4dyg}. The advancements in dynamic graph neural networks have made it possible to learn evolutionary patterns from non-stationary temporal data~\cite{DBLP:conf/kdd/KumarZL19}. However, effectively capturing the multi-scale temporal dependencies in network dynamics remains an open challenge.
	
	Dynamic link prediction faces fundamental challenges due to the multi-scale and complex temporal patterns in network evolution~\cite{DBLP:journals/corr/abs-2006-10637,DBLP:conf/sigir/0001GRTY20}. These patterns include periodic fluctuations in node interactions, long-range temporal dependencies influenced by global network states, and abrupt topological changes caused by sudden events~\cite{weisfeiler1968reduction,aamand2022exponentially}. To illustrate this point, consider the context of academic collaborative networks, two researchers may establish collaborations due to regular participation in periodic academic events, while their collaboration intensity is also influenced by long-term factors, such as the development trends in their research field. Existing methods frequently fail to model such cross-scale dependencies, resulting in systematic biases in predictions. For instance, temporary interaction pauses may be misinterpreted as permanent collaboration termination, rather than recognizing recurring patterns tied to periodic events.
	
	The prevailing methodologies are encumbered by significant limitations. Firstly, architectures based on recurrent neural networks~\cite{DBLP:journals/corr/abs-2006-10637}, while capable of handling sequential data, struggle to capture long-range dependencies due to their serial computation nature, with the root cause being gradient vanishing or explosion, which leads to the loss of long-distance dependencies~\cite{DBLP:conf/sigmod/WangLLXYWWCYSG21,luo2022neighborhoodaware}. Secondly, conventional temporal attention mechanisms, e.g., TGAT and DyGFormer ~\cite{DBLP:conf/iclr/XuRKKA20,yu2023towards} are deficient in their ability to differentiate between disparate frequency patterns in nonstationary time intervals, particularly in networks characterized by pronounced periodicity or sudden fluctuations. Thirdly, purely frequency domain methods, while effective in capturing global frequency characteristics, are ineffective in capturing localized temporal details, rendering them unsuitable for nonstationary signals. Finally, it has been demonstrated that fixed window-length mechanisms are incapable of adapting to variable-period patterns, thus limiting their applicability to real-world dynamic networks with evolving cycles.

	To address the aforementioned bottlenecks, it is essential to break through the limitations of traditional algorithmic frameworks. In response, this study proposes the Temporal-Frequency Collaborative Wavelet-Transformer (TFWaveFormer) framework to address the problem of multi-scale temporal modeling in dynamic link prediction through a temporal-frequency collaborative approach. The time domain excels at capturing local dynamics, directly modeling the microscopic evolution of nodes and edges across consecutive timestamps. For instance, TGAT accurately captures instantaneous interaction patterns in social networks through temporal attention, making event stream modeling more natural~\cite{velivckovic2017graph}. On the other hand, wavelet transformation achieves precise extraction of local temporal-frequency features in dynamic graph data through multi-resolution analysis, overcoming the global averaging limitations of traditional frequency-domain methods. Its adaptive temporal-frequency windowing is particularly well-suited for analyzing non-stationary graph signals. By integrating temporal-frequency collaborative mechanisms, the proposed framework achieves effective decoupling of long-term evolutionary trends and instantaneous dynamic patterns through gated fusion and cross-scale interaction, simultaneously addressing the challenges of frequency pattern differentiation, localized temporal detail preservation, and adaptive cycle adaptation in non-stationary dynamic networks.
	
	From an implementation perspective, TFWaveFormer adopts a three-stage processing workflow. First, positional encoding is introduced on the integrated features, and the Transformer architecture is utilized for multi-scale decomposition and interaction modeling. Subsequently, a WaveletFilter is designed to simulate wavelet transforms through multi-scale convolution kernels, extracting multi-scale dynamic features from temporal sequences. The wavelet-domain features are coupled with the multi-head attention mechanism of the Transformer, further capturing the complex dynamic relationships between nodes and their neighborhoods.
	Finally, a novel attention-based temporal-frequency fusion framework is proposed, effectively integrating temporal and frequency-domain features to aggregate the features of nodes and generate the final embeddings for source and target nodes. This design not only retains the parallel computation advantages of standard Transformers but also enhances the model's ability to identify complex temporal patterns through frequency-domain analysis.
	Our contributions can be summarized as follows:
	\begin{itemize}[itemsep=2pt,topsep=-2pt,parsep=0pt,leftmargin=10pt]
		\item We propose a novel learnable wavelet decomposition module that replaces traditional fixed-basis transforms with parallel, multi-scale convolutional kernels. This allows for the adaptive and efficient extraction of both fine-grained local and coarse-grained global temporal patterns directly from the data.
		\item We design a temporal-frequency coordination mechanism that synergistically integrates representations from both time and frequency domains to further effectively capture complementary dynamics, ranging from transient events to long-term periodicities in link evolution.
		\item We  validate TFWaveFormer, a unified Transformer architecture that seamlessly integrates our proposed wavelet and co-attention modules. Extensive experiments demonstrate that TFWaveFormer establishes a new state-of-the-art in dynamic link prediction, significantly outperforming existing methods across ten benchmark datasets of varying scales.
	\end{itemize}

	\section{Related Work}
	Dynamic graphs have garnered significant attention for their ability to model evolving relationships in real-world systems~\cite{wang2024large,DBLP:conf/iclr/XuRKKA20,Shi_Fan_Kwok_2020}. Early methods extended static models with recurrent units~\cite{DBLP:conf/nips/SouzaMKG22} or time-aware attention~\cite{DBLP:conf/iclr/XuRKKA20}. Subsequent approaches introduced specialized temporal architectures: JODIE~\cite{DBLP:conf/kdd/KumarZL19} uses coupled RNNs for interaction dynamics, while TGAT~\cite{DBLP:conf/iclr/XuRKKA20} and DySAT~\cite{DBLP:conf/wsdm/SankarWGZY20} leverage temporal self-attention. More recent models, such as GraphMixer~\cite{cong2023do}, adopt MLPs and pooling to simplify temporal modeling, while TGN~\cite{DBLP:journals/corr/abs-2006-10637} introduces a memory-based continuous-time framework. CAWN~\cite{DBLP:conf/iclr/WangCLL021} exploits anonymized temporal walks, and Transformer-based models like DyGFormer~\cite{yu2023towards} and CorDGT~\cite{wang2025dynamic} enhance long-range temporal representation and spatiotemporal locality, respectively.
	
	Despite these advances, most temporal models overlook frequency-domain representations, which can provide complementary insights into dynamic behaviors. Several studies have applied spectral filtering for tasks such as anomaly detection and link prediction~\cite{xu2024revisiting,10.1145/3583780.3615067}. Spectral decomposition captures frequency components of graph signals~\cite{xiong2025spectral}, and wavelet transforms support multiresolution analysis of non-stationary dynamics~\cite{shalby2025comprehensive,gao2024efficient}. FFT-based methods like FreeDyG~\cite{tian2023freedyg} are effective at extracting periodic patterns.
	However, existing models often fail to jointly capture micro-level temporal events and macro-level structural trends. This gap calls for a unified framework that integrates temporal and frequency perspectives via multi-scale modeling.

	\begin{figure*}[h]
		\centering
		\includegraphics[scale=0.55]{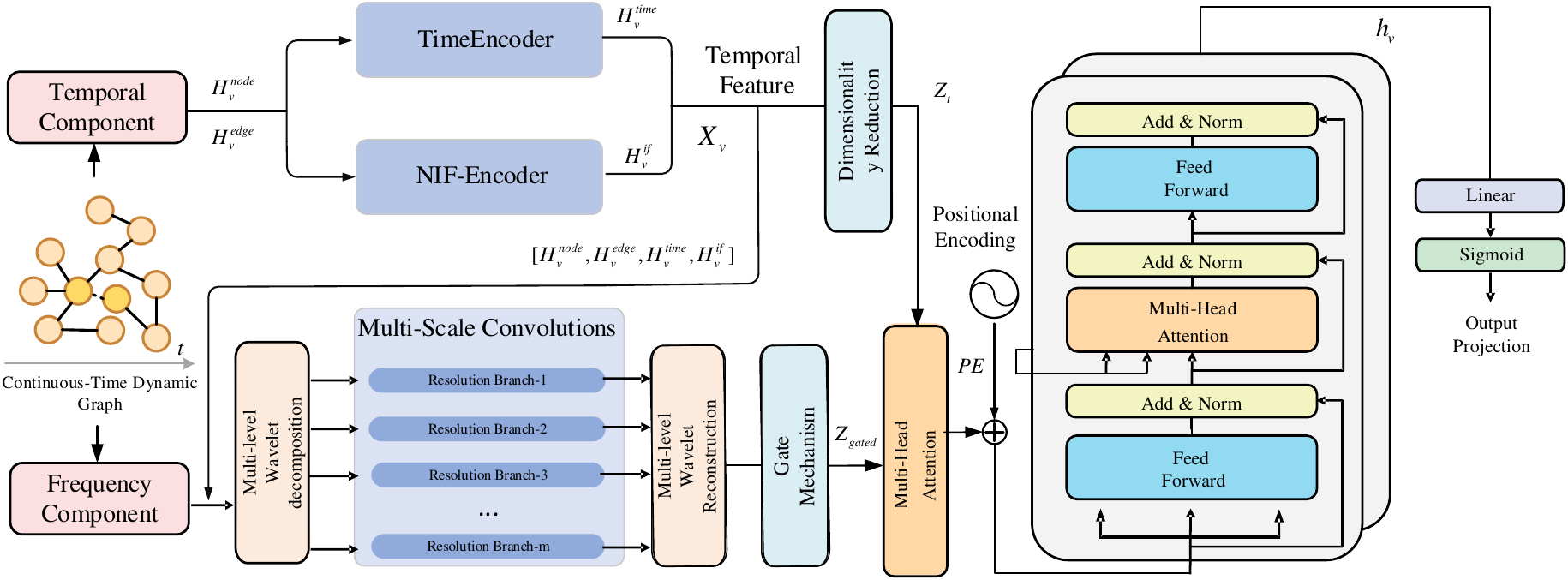}
		\caption{The proposed TFWaveFormer framework. The architecture consists of three key components: (1) feature integration, (2) multi-level wavelet transformation, and (3) temporal-frequency hybrid transformer for generating final representations.}
		\label{4}
	\end{figure*}
	\section{Preliminary}
	%This section establishes the fundamental concepts and mathematical formulations underlying our TFWaveFormer approach for dynamic graph learning. 
	We begin by formally defining the dynamic graph representation problem and then introduce the key theoretical foundations that motivate our architectural design.
	
	\textbf{Problem Formulation:}
	This work considers a dynamic graph as a sequence of non-decreasing chronological interactions $\mathbf{X}=\left\{\left(u_1,v_1,t_1\right), \left(u_2,v_2,t_2\right), \cdots \right\}$ with $0 \leq t_1 \leq t_2$, where $u_i, v_i \in \mathcal{V}$ denote the source node and destination node of the $i$-th edge at timestamp $t_i$. %$\mathbf{X}_v$ represents the historical interactions associated with node $v$. 
	And node $v$'s neighbors are nodes connected to $v$:
	%	\begin{equation}
		%		\mathcal N_t(v) = \{ u|(u,v,t') \in \mathcal E,t' \le t\},
		%	\end{equation}
	%	where $\mathcal{E}$ represents the edge set of the dynamic graph. 
	The purpose of this algorithm is to effectively learn and represent the feature embeddings of nodes by leveraging both the target nodes themselves and their historical neighbors, and predict the probability of a link existing between target nodes.

	\textbf{Wavelet Transform Foundations:}
	Traditional wavelet transforms provide a mathematical framework for multi-resolution analysis through decomposition of signals into different frequency components. Given $x(t)$, the continuous wavelet transform is defined as:
	\begin{equation}
		W(a,b) = \frac{1}{\sqrt{a}} \int_{-\infty}^{\infty} x(t) \psi^*\left(\frac{t-b}{a}\right) dt,
	\end{equation}
	where $\psi(t)$ is the mother wavelet, $a$ is the scale parameter, $b$ is the translation parameter, and $\psi^*$ denotes the complex conjugate.
	
	In the temporal setting relevant to dynamic graphs, we adapt this principle to analyze temporal interaction patterns at multiple scales simultaneously. Unlike traditional approaches that rely on fixed wavelet bases, our method employs learnable convolutional kernels $\Psi = \{\psi_k^{(i)} \in \mathbb{R}^{k} | k \in \mathcal{K}, i \in [1,d]\}$ to achieve data-driven wavelet-like decomposition, enabling adaptive discovery of optimal temporal patterns for dynamic graph analysis.

	\section{METHODOLOGY}
	
	%The TFWaveFormer architecture we proposed that consists of three key components: Multi-level Wavelet Transformation, Gate Mechanism, and Temporal-Frequency Feature Fusion as shown in Fig.1. These components effectively capture both the temporal and frequency-domain features of dynamic graph data.
	The TFWaveFormer architecture consists of three key components: \textbf{feature integration}, \textbf{multi-level wavelet} and \textbf{temporal-frequency hybrid transformer}. As shown in Figure 1, the original data first undergoes feature integration, then dimension compression and multi-level wavelet transformation respectively, and finally the final representation is achieved through temporal-frequency hybrid transformer. Algorithm 1 presents the overall workflow of the proposed algorithm for better understanding. %The latter two are our focus, which we focus on in this section. Next, the multi-level wavelet transformation module is introduced in detail.
	\subsection{Feature Extraction}
	Dynamic graphs encode information through multiple modalities. For each node $v$ in the prediction task, we extract historical interaction patterns to construct rich feature representations.
	The node feature ${H}^{node}_v \in \mathbb{R}^{d_n}$ contains inherent attributes of neighbor nodes, where $d_n$ is the node feature dimension. The edge feature ${H}^{edge}_v \in \mathbb{R}^{d_e}$ captures interaction semantics, where $d_e$ represents edge feature dimension.
	Temporal features are processed through a time encoder:
	\begin{equation}
		{H}^{time}_v = \text{Time-Encoder}([t - t_1, t - t_2, ..., t - t_L]),
	\end{equation}
	where $t$ is the current timestamp and $t_i$ represents historical interaction times.
	Node Interaction frequency (NIF) features capture topological relationships:
	\begin{equation}
		\begin{split}
			{H}^{if}_v = \text{NIF-Encoder}([\text{count}(\mathcal{N}(v) \cap \mathcal{N}(u)),
			\text{freq}(\mathcal{N}(v))])
		\end{split}
	\end{equation}
	where $\mathcal{N}(v)$ denotes the neighbor set of node $v$.
	To unify different modalities, feature alignment maps all features to a common space:
	\begin{equation}
		{F}^{(m)}_v = {H}^{(m)}_v \mathbf{W}^{(m)} + \mathbf{b}^{(m)},
	\end{equation}
	where $m \in \{\text{node}, \text{edge}, \text{time}, \text{if}\}$, and $\mathbf{W}^{(m)}$, $\mathbf{b}^{(m)}$ are learnable parameters.
	Final fused features are obtained through concatenation-compression:
	\begin{equation}
		\begin{split}
			\mathbf{X}_v = \text{ReduceLayer}(\text{Concat}([{F}^{(node)}_v, \\
			{F}^{(edge)}_v, {F}^{(time)}_v, {F}^{(if)}_v])),
		\end{split}
	\end{equation}
	\noindent
	where $\mathbf{X}_v$ denotes the feature representation of node $v$, serving as the input for subsequent wavelet transformation and feature extraction.
	\subsection{Multi-Level Wavelet}
	Given the input features $\mathbf{X}_v \in \mathbb{R}^{L \times d}$, where $L$ represents the length of historical interactions, and $d$ indicates the dimension. We define the temporal domain as $\mathcal{T} = \{t_1, t_2, ..., t_L\}$, and employ multi-level wavelet transform to decompose dynamic graph data. Our approach inherits the multi-resolution analysis principle of traditional wavelet transforms while adopting a parallel multi-scale convolutional architecture to achieve more flexible and efficient temporal-frequency decomposition. Unlike traditional methods that utilize predefined wavelet basis functions such as Daubechies and Symlets~\cite{necula2012transient}, our method realizes data-driven adaptive wavelet decomposition through learnable multi-scale convolutional kernels $\Psi = \{\psi_k^{(i)}\}_{i=1}^{d}$.
	\subsubsection{Decomposition:}
	The multi-level wavelet transform leverages a series of learnable convolutional filters $\psi_k^{(i)}$ with different scales $k$ to simulate wavelet-like behavior and extract multi-scale temporal patterns. The wavelet transformation is applied independently to each feature dimension.
	We define the scale space as $\mathcal{K} = \{k_1, k_2, ..., k_m\}$. In this context, the wavelet transform can be formulated as:
%	\begin{equation}
%		\mathbf{Z}_{k}^{(i)} = \sum_{j=1}^{L} \mathbf{X}_v^{(j)} \ast \psi_k^{(i,j)}, \quad \forall k \in \mathcal{K}, \forall i \in [1,d],
%	\end{equation}
	\begin{equation}
		\mathbf{Z}_{k}^{(i)}[t] = \sum_{j \in \mathcal{R}_k} \psi_k^{(i)}[j] \cdot \mathbf{X}_v^{(i)}[t+j], \quad \forall k \in \mathcal{K}, \forall i \in [1,d],
	\end{equation}
	\begin{equation}
		\begin{aligned}
			\mathcal{Z} &= \{Z_{k_1}^{(1:d)}, Z_{k_2}^{(1:d)}, Z_{k_3}^{(1:d)}, ...,  Z_{k_m}^{(1:d)}\}\\
			&	=\text{Decomp}(\mathbf{X}_v, \Psi, m, \theta),
		\end{aligned}
	\end{equation}
	where $\ast$ denotes the convolution operation, $\psi_k^{(i)}[j]$ represents the weight at position $j$ of the convolutional kernel corresponding to scale $k$ and channel $i$, and $\theta \in \Theta$ denotes the learnable parameter space. Each $\psi_k^{(i)}$ is parameterized as a depth-wise separable convolutional filter with kernel size $k$ to capture temporal dependencies at different granularities. We define the temporal receptive field as $\mathcal{R}_k = \{r | |r| \leq \lfloor k/2 \rfloor\}$. For instance, smaller kernel sizes correspond to fine-grained short-term patterns, while larger kernel sizes capture coarse-grained long-term trends. This approach ensures effective modeling of both local and global temporal structures. Here, $m$ denotes the number of wavelet convolutional kernels in the wavelet decomposition. We only retain the effective channels from the final level $m$, as excessive convolutional kernels do not yield significant performance improvements but rather increase computational overhead.

	\subsubsection{Parallel Representation:}	
	The iterative decomposition process of the traditional wavelet transform is replaced by parallelized multi-scale convolution, the convolutional filters $\Psi = \{\psi_k^{(i)} \in \mathbb{R}^{k} | k \in \mathcal{K}, i \in [1,d]\}$ are fully learnable, enabling the model to adaptively discover the most relevant temporal patterns from data. The parameters are updated by optimizing the objective function $\mathcal{L}(\theta)$:
	\begin{equation}
		\theta^* = \arg\min_{\theta \in \Theta} \mathcal{L}(\theta) + \lambda \|\Psi\|^2,
	\end{equation}
	where $\lambda$ is the regularization coefficient. These filters operate in a depth-wise separable manner, with each feature channel $i$ processed independently:
	\begin{equation}
		\mathcal{C}_k^{(i)}(\mathbf{X}_v) = \sum_{j \in \mathcal{R}_k} \psi_k^{(i,j)} \cdot \mathbf{X}_v^{(t+j,i)}, \quad \forall t \in [1,L].
	\end{equation}
	
	Significantly reducing computational costs while maintaining expressive power. The convolutional layers are appropriately padded with symmetric padding strategy $\mathcal{P}_k = \lfloor k/2 \rfloor$ for odd kernel sizes and asymmetric padding for even kernel sizes to ensure all outputs $\mathbf{Z}_k$ maintain the same temporal resolution $L$ as the input.

	%\subsubsection{Multi-scale Wavelet Representation:}
	The outputs of convolutional filters $\mathbf{Z}_k \in \mathbb{R}^{L \times d}$ at each scale are further combined into a unified multi-scale representation. We define the scale attention function $\mathcal{A}: \mathbb{R}^{|\mathcal{K}| \times d} \rightarrow \mathbb{R}^{|\mathcal{K}| \times d}$. To effectively integrate cross-scale information, we introduce learnable scale weights $\mathbf{W} = \{\mathbf{w}_k \in \mathbb{R}^d | k \in \mathcal{K}\}$. These weights are normalized through a softmax function $\varsigma$ to ensure stable contributions across different scales:
	\begin{equation}
		\mathbf{S}_k = \frac{\exp(\mathbf{w}_k / \tau)}{\sum_{k' \in \mathcal{K}} \exp(\mathbf{w}_{k'} / \tau)}, \quad \tau > 0
	\end{equation}
	where $\tau$ is the temperature parameter. The resulting multi-scale wavelet representation $\mathbf{Z}_{ms}\in\mathbb{R}^{L \times d}$ is computed as follows:
	\begin{equation}
		\mathbf{Z}_{ms} = \sum_{k \in \mathcal{K}} \mathbf{S}_k \odot \mathbf{Z}_k, %= \sum_{i=1}^{m} \alpha_i \cdot \mathcal{C}_{k_i}(\mathbf{X}_v),
	\end{equation}
	where $\odot$ denotes the hadamard product. This mechanism enables the model to dynamically attend to temporal patterns most relevant to the given task.
	
	\subsubsection{Reconstruction:}
	To adaptively assess feature importance across time and channels, we introduce a gating mechanism $\mathcal{G}\in \mathbb{R}^{L \times d}$ that generates a gate vector for each temporal step and feature dimension. We define two MLP networks $f_1, f_2: \mathbb{R}^d \rightarrow \mathbb{R}^d$. The gate mechanism is computed as:
	\begin{equation}
		\mathcal{G} = \sigma\left(f_2\left(\text{GELU}(f_1(\mathbf{Z}_{ms}) + \mathbf{b}_1)\right) + \mathbf{b}_2\right),
	\end{equation}
	where $\sigma(\cdot)$ denotes the sigmoid function, $\text{GELU}(\cdot)$ represents the Gaussian Error Linear Unit, and $\mathbf{b}_1, \mathbf{b}_2 \in \mathbb{R}^d$ are bias terms. The gate $\mathcal{G}$ modulates the multi-scale features through element-wise multiplication, thereby achieving the reconstruction of multi-level wavelet transform:
	\begin{equation}
		\mathbf{Z}_{gated} = \mathcal{G} \odot \mathbf{Z}_{ms}. %= \sum_{j=1}^{L} \sum_{k=1}^{d} g_{jk} \cdot z_{jk}.
	\end{equation}
	Subsequently, the refined representation $\mathbf{Z}_{gated} \in \mathbb{R}^{L \times d}$ is passed to the temporal-frequency fusion module for integration with temporal-domain features, enabling comprehensive dynamic representation learning.
	\begin{algorithm}[H]
		\caption{TFWaveFormer Algorithm}
		\label{alg:tfwaveformer}
		\begin{algorithmic}[1]
			\REQUIRE Node $v$, length $L$, scales $\mathcal{K}$, heads $h$
			\ENSURE Node embedding $\hat{\mathbf{h}}_v$, prediction $\hat{y}_{uv}$
			
			\STATE \textbf{Feature Integration:}
			\STATE $\mathbf{X}_v \leftarrow \text{Fuse}([{H}^{node}_v, {H}^{edge}_v, {H}^{time}_v, {H}^{if}_v])$
			
			\STATE \textbf{Multi-Level Wavelet:}
			\STATE \textit{// Decomposition:}
			\FOR{$k \in \mathcal{K}$, $i \in [1,d]$, $t \in [1,L]$}
			\STATE $\mathbf{Z}_{k}^{(i)}[t] \leftarrow \sum_{j \in \mathcal{R}_k} \psi_k^{(i)}[j] \cdot \mathbf{X}_v^{(i)}[t+j]$
			\ENDFOR
			
			\STATE \textit{// Parallel Representation:}
			\STATE $\mathbf{S}_k \leftarrow \text{Softmax}(\mathbf{w}_k / \tau)$ for all $k \in \mathcal{K}$
			\STATE $\mathbf{Z}_{ms} \leftarrow \sum_{k \in \mathcal{K}} \mathbf{S}_k \odot \mathbf{Z}_k$
			
			\STATE \textit{// Reconstruction:}
			\STATE $\mathcal{G} \leftarrow \sigma(f_2(\text{GELU}(f_1(\mathbf{Z}_{ms})+\mathbf{b}_1))+\mathbf{b}_2)$
			\STATE $\mathbf{Z}_{gated} \leftarrow \mathcal{G} \odot \mathbf{Z}_{ms}$
			
			\STATE \textbf{Temporal-Frequency Hybrid Transformer:}
			\STATE $\mathbf{Z}^{0} \leftarrow \text{LayerNorm}(\text{MLP}(\mathbf{X}_v) + \mathbf{Z}_{gated} + \text{PE})$
			\STATE $\mathbf{Z}^{1} \leftarrow \text{LayerNorm}(\mathbf{Z}^{0} + \text{MHSA}(\mathbf{Z}^{0}))$
			\STATE $\mathbf{h}_v \leftarrow \text{LayerNorm}(\mathbf{Z}^{1} + \text{MLP}(\mathbf{Z}^{1}))$
			
			\STATE \textbf{Dynamic Link Prediction:}
			\STATE $\hat{\mathbf{h}}_v \leftarrow \frac{1}{L}\sum_{t=1}^{L}\mathbf{h}_v[t]$
			\STATE $s_{uv} \leftarrow \mathbf{w}^T (\hat{\mathbf{h}}_u \odot \hat{\mathbf{h}}_v) + b$
			\STATE $\hat{y}_{uv} \leftarrow \sigma(s_{uv})$
			
			\RETURN $\hat{\mathbf{h}}_v$, $\hat{y}_{uv}$
		\end{algorithmic}
	\end{algorithm}
	\subsection{Temporal-Frequency Hybrid Transformer}
	The third step is to fuse temporal-frequency features based on the multi-head self-attention mechanism. Firstly, the position encoding $PE$ as follows:
	\begin{equation}
		\begin{split}
			PE(pos, 2i) &= \sin(pos/10000^{2i/d}), \\
			PE(pos, 2i+1) &= \cos(pos/10000^{2i/d}),
		\end{split}
	\end{equation}
	where $pos$ is the position index and $i$ is the dimension index. The temporal feature is compressed by MLP encoding $\mathbf{Z}_{t} = MLP(\mathbf{X}_v)$ and combined with the final output of the frequency domain feature $\mathbf{Z}_{{gated}}$for self-attention fusion:
	\begin{equation}
		\mathbf{Z}^{0} = \text{LayerNorm}(\mathbf{Z}_t + \mathbf{Z}_{gated}+PE),
	\end{equation}
	\begin{equation}
		\mathbf{Z}^{1} = \text{LayerNorm}(\mathbf{Z}^{0} + \text{MHSA}(\mathbf{Z}^{0})),
	\end{equation}
	\begin{equation}
		\mathbf{h}_v = \text{LayerNorm}(\mathbf{Z}^{1} + MLP(\mathbf{Z}^{1})),
	\end{equation}
	where $\text{LayerNorm}(\cdot)$ is standard normalization. $\text{MHSA}(\cdot)$ denotes the multi-head self-attention mechanism with $h$ parallel heads:
	\begin{equation}
		\text{MHSA}(\mathbf{Z}) = \text{Concat}(\text{head}_1, ..., \text{head}_h)\mathbf{W}^O,
	\end{equation}
	\begin{equation}
		\text{head}_i = \text{Softmax}\left(\frac{\mathbf{Q}_i\mathbf{K}_i^\top}{\sqrt{d/h}}\right)\mathbf{V}_i,
	\end{equation}	
	where $\mathbf{Q}_i = \mathbf{Z}\mathbf{W}_i^Q$, $\mathbf{K}_i = \mathbf{Z}\mathbf{W}_i^K$, $\mathbf{V}_i = \mathbf{Z}\mathbf{W}_i^V$ are the query, key, and value projections with $\mathbf{W}_i^Q, \mathbf{W}_i^K, \mathbf{W}_i^V \in \mathbb{R}^{d \times (d/h)}$, and $\mathbf{W}^O \in \mathbb{R}^{d \times d}$ is the output projection matrix. The complete transformation can be formalized as TFWaveFormer Layer with the output $\mathbf{h}_v$. Note that this represents only a single layer within our model, the full architecture comprises at least two stacked layers to ensure sufficient representational capacity.
	\subsection{Dynamic Link Prediction}
	After training, the learned model performs dynamic link prediction. Based on the nodes $u$ and $v$ embedding representations $\mathbf{h}_u$ and $\mathbf{h}_v$, the dynamic link prediction module generates scores. We first apply temporal aggregation to obtain fixed-size node representations from the sequence outputs $\mathbf{h}_v \in \mathbb{R}^{L \times d}$ using mean pooling: $\hat{\mathbf{h}}_v = \frac{1}{L}\sum_{t=1}^{L}\mathbf{h}_v[t]$. The prediction score is computed as:
	\begin{equation}
		s_{uv} = \mathbf{w}^T (\hat{\mathbf{h}}_u \odot \hat{\mathbf{h}}_v) + b,
	\end{equation}
	\begin{equation}
		\hat{y}_{uv} = \sigma(s_{uv}),
	\end{equation}
	where $\mathbf{w}$ and $b$ are learnable parameters, $s_{uv}$ is the pre-sigmoid score, and $\hat{y}_{uv} \in [0,1]$ represents the predicted link formation probability. Model training employs the cross-entropy loss function:
	\begin{equation}
		\mathcal{L} = \frac{1}{|\mathcal{E}|} \sum_{(u,v) \in \mathcal{E}} \log(1 + \exp(-y_{uv}^* \cdot s_{uv})),
	\end{equation}
	where $\mathcal{E}$ denotes the set of all candidate node pairs in the training set and $y_{uv}^* = 2y_{uv} - 1 \in \{-1, 1\}$ represents the transformed ground truth labels. This design achieves end-to-end learning from node representations to link prediction probabilities.
	\section{Experimental Design and Evaluation}
	\subsection{Datasets and benchmarks}
	%This research carries out experimental evaluations on eight cross-domain datasets, spanning various domains including social networks (Wikipedia, Reddit), online education (MOOC), Social Evolution(Social Evo.), user behavior (LastFM), communication (Enron), mobility trajectories (UCI), Flights, Contact and UN Trade \cite{yu2023towards}. 
	%All experiments are conducted on a Intel(R) Xeon(R) Gold 6326 CPU @ 2.90GHz, and NVIDIA RTX A6000, with implementations based on PyTorch. The proposed TFWaveFormer model is benchmarked against cutting-edge dynamic graph learning methods, which incorporate diverse approaches: DyRep~\cite{DBLP:conf/iclr/TrivediFBZ19}, TGN~\cite{DBLP:journals/corr/abs-2006-10637}, CAWN~\cite{DBLP:conf/iclr/WangCLL021}, and GraphMixer~\cite{cong2023do}, DyGFormer~\cite{yu2023towards}, and FreeDyG~\cite{tian2023freedyg}, CorDGT~\cite{wang2025dynamic}. 
	All experiments are conducted on an Intel(R) Xeon(R) Gold 6326 CPU @ 2.90GHz and NVIDIA RTX A6000. The proposed TFWaveFormer model is benchmarked against cutting-edge dynamic graph learning methods, which incorporate diverse approaches: DyRep~\cite{DBLP:conf/iclr/TrivediFBZ19}, TGN~\cite{DBLP:journals/corr/abs-2006-10637}, CAWN~\cite{DBLP:conf/iclr/WangCLL021}, as well as recently proposed methods including GraphMixer~\cite{cong2023do}, DyGFormer~\cite{yu2023towards}, FreeDyG~\cite{tian2023freedyg}, CorDGT~\cite{wang2025dynamic}, CTAN~\cite{gravina2024long}, and DyGMamba~\cite{ding2024dygmamba}.
	
	We evaluate the proposed method across ten real-world datasets including Wikipedia, Reddit, MOOC, Social Evo., LastFM, Enron, UCI, Flights, Contact, and UN Trade, covering a broad spectrum of applications such as online collaboration, social interaction, education, communication, mobility, and international trade~\cite{yu2023towards}. As shown in Table 4 of Appendix A.1, these include both small-scale (Enron, UN Trade, Contact, UCI) and large-scale (LastFM, MOOC, Wikipedia, Reddit, Flights, Social Evo.) networks. More detailed can be found in Appendix A.1. 
	Performance is assessed using Average Precision (AP) and Area Under the ROC Curve (AUC), with datasets chronologically split into 70\% training, 15\% validation, and 15\% testing. 
\begin{table*}[h] 
	\centering
	\caption{Results for transductive dynamic link prediction with Average Precision and Area Under the ROC metrics under random negative sampling strategy.}
	%\footnotesize  % 符合AAAI 9号字体要求
	%\setlength{\tabcolsep}{2pt}  % 压缩列间距
	\resizebox{1.0\textwidth}{!}{
		\setlength{\tabcolsep}{0.9mm}{
			\begin{tabular}{c|c|cccccccccc}
				\toprule
				Metr.\rule{0pt}{9pt} & Datasets & DyRep & TGN & CAWN & GraphMixer & FreeDyG & DyGFormer & CorDGT & CTAN & DyGMamba & OURS \\ \hline
				\multirow{10}{*}{AP}  
				& Wikipedia  & 94.86 $\pm$ 0.06 & 98.45 $\pm$ 0.06 & 98.76 $\pm$ 0.03 & 97.25 $\pm$ 0.03 & {\underline{99.23 $\pm$ 0.01}} & 98.82 $\pm$ 0.02 & 99.07 $\pm$ 0.01 & 96.61 $\pm$ 0.79 & 99.15 $\pm$ 0.02 & {\textbf{99.33 $\pm$ 0.01}}  \\
				& Reddit     & 98.22 $\pm$ 0.04 & 98.63 $\pm$ 0.06 & 99.11 $\pm$ 0.01 & 97.31 $\pm$ 0.01 & 99.21 $\pm$ 0.01 & 99.11 $\pm$ 0.01 & 99.18 $\pm$ 0.00 & 97.21 $\pm$ 0.84 & {\underline{99.31 $\pm$ 0.01}} & {\textbf{99.32 $\pm$ 0.01}} \\ 
				& MOOC       & 81.97 $\pm$ 0.49 & 89.15 $\pm$ 1.60 & 80.15 $\pm$ 0.25 & 82.78 $\pm$ 0.15 & 87.23 $\pm$ 0.45 & 89.01 $\pm$ 0.11 & 84.35 $\pm$ 0.14 & 84.71 $\pm$ 2.85 & \underline{89.21 $\pm$ 0.08} & \textbf{91.24 $\pm$ 0.05} \\ 
				& LastFM     & 71.92 $\pm$ 2.21 & 77.07 $\pm$ 3.97 & 86.99 $\pm$ 0.06 & 75.61 $\pm$ 0.24 & 90.74 $\pm$ 0.14 & 93.00 $\pm$ 0.12 & 92.23 $\pm$ 0.10 & 86.44 $\pm$ 0.80 & \underline{93.35 $\pm$ 0.20} & \textbf{94.64 $\pm$ 0.05} \\ 
				& Enron      & 82.38 $\pm$ 3.36 & 86.53 $\pm$ 1.11 & 89.56 $\pm$ 0.09 & 82.25 $\pm$ 0.16 & 92.51 $\pm$ 0.05 & 92.47 $\pm$ 0.12 & 91.76 $\pm$ 0.50 & 92.52 $\pm$ 1.20 & \underline{92.65 $\pm$ 0.12} & \textbf{92.70 $\pm$ 0.20} \\ 
				& Social Evo.& 88.87 $\pm$ 0.30 & 93.57 $\pm$ 0.17 & 84.96 $\pm$ 0.09 & 93.37 $\pm$ 0.07 & 94.54 $\pm$ 0.01 & {94.53 $\pm$ 0.01} & 93.81 $\pm$ 0.02 & Timeout & \textbf{94.77 $\pm$ 0.01} & \underline{94.65 $\pm$ 0.01} \\ 
				& UCI        & 65.14 $\pm$ 2.30 & 92.34 $\pm$ 1.04 & 95.18 $\pm$ 0.06 & 93.25 $\pm$ 0.57 & {{\underline{96.03 $\pm$ 0.02}}} & 95.76 $\pm$ 0.17 & 96.03 $\pm$ 0.02 & 76.64 $\pm$ 4.11 & 95.91 $\pm$ 0.15 & \textbf{96.51 $\pm$ 0.03} \\ 
				& Flights    & 95.29 $\pm$ 0.72 & 97.95 $\pm$ 0.14 & 98.51 $\pm$ 0.01 & 90.99 $\pm$ 0.05 & 98.20 $\pm$ 0.01 & 98.25 $\pm$ 0.01 & {{\underline{98.81 $\pm$ 0.01}}} & Timeout & 98.12$\pm$ 0.01 & \textbf{98.87 $\pm$ 0.01} \\ 
				& UN Trade   & 63.21 $\pm$ 0.93 & 65.03 $\pm$ 1.37 & {{\underline{65.39 $\pm$ 0.12}}} & 62.61 $\pm$ 0.27 & 50.00 $\pm$ 0.01 & 64.25 $\pm$ 0.43 & 50.00 $\pm$ 0.01 & 50.01$\pm$ 0.01 & 64.58$\pm$ 0.22 & \textbf{66.06 $\pm$ 0.24} \\ 
				& Contact    & 95.98 $\pm$ 0.15 & 96.89 $\pm$ 0.56 & 90.26 $\pm$ 0.28 & 91.92 $\pm$ 0.03 & 98.00 $\pm$ 0.01 & {{\textbf{98.29 $\pm$ 0.01}}} & 97.79 $\pm$ 0.04 & Timeout & 97.71$\pm$ 0.03 & {\underline{98.12 $\pm$ 0.01}} \\  
				\cline{2-12}
				& Avg. Rank \rule{0pt}{8pt} & 8.50&6.30&6.30&8.20&4.10&4.20&4.90&6.40&2.40&\textbf{1.20}\\ \hline
				\multirow{10}{*}{AUC}  
				& Wikipedia  & 94.37 $\pm$ 0.09 & 98.37 $\pm$ 0.07 & 98.54 $\pm$ 0.04 & 96.92 $\pm$ 0.03 & \underline{99.20 $\pm$ 0.01} & 98.91 $\pm$ 0.02 & 98.84 $\pm$ 0.01 & 97.00 $\pm$ 0.21 & 99.08 $\pm$ 0.02 & \textbf{99.31 $\pm$ 0.01} \\ 
				& Reddit     & 98.17 $\pm$ 0.05 & 98.60 $\pm$ 0.06 & 99.01 $\pm$ 0.01 & 97.17 $\pm$ 0.02 & 99.19 $\pm$ 0.01 & 99.15 $\pm$ 0.01 & 99.12 $\pm$ 0.01 & 97.24 $\pm$ 0.75 & \underline{99.27 $\pm$ 0.01} & \textbf{99.29 $\pm$ 0.01} \\ 
				& MOOC       & 85.03 $\pm$ 0.58 & \underline{91.21 $\pm$ 1.15} & 80.38 $\pm$ 0.26 & 84.01 $\pm$ 0.17 & 89.12 $\pm$ 0.39 & 87.91 $\pm$ 0.58 & 85.78 $\pm$ 0.61 & 85.40 $\pm$ 2.67 & 89.58 $\pm$ 0.12 & \textbf{91.89 $\pm$ 0.35} \\
				& LastFM     & 71.16 $\pm$ 1.89 & 78.47 $\pm$ 2.94 & 85.92 $\pm$ 0.10 & 73.53 $\pm$ 0.12 & 89.56 $\pm$ 0.12 & 93.05 $\pm$ 0.10 & 92.37 $\pm$ 0.02 & 85.12 $\pm$ 0.77 & \underline{93.31 $\pm$ 0.18} & \textbf{94.52 $\pm$ 0.04} \\ 
				& Enron      & 84.89 $\pm$ 3.00 & 88.32 $\pm$ 0.99 & 90.45 $\pm$ 0.14 & 84.38 $\pm$ 0.21 & \underline{93.81 $\pm$ 0.11} & {93.33 $\pm$ 0.13} & 93.28 $\pm$ 0.04 & 87.09 $\pm$ 1.51 & 93.34 $\pm$ 0.23 & \textbf{93.97 $\pm$ 0.09} \\ 
				& Social Evo.& 90.76 $\pm$ 0.21 & 95.39 $\pm$ 0.17 & 87.34 $\pm$ 0.08 & 95.23 $\pm$ 0.07 & 96.32 $\pm$ 0.01 & 96.30 $\pm$ 0.01 & 95.48 $\pm$ 0.01 & Timeout & \textbf{96.38 $\pm$ 0.02} & \underline{96.38 $\pm$ 0.02} \\ 
				& UCI        & 68.77 $\pm$ 2.34 & 92.03 $\pm$ 1.13 & 93.87 $\pm$ 0.08 & 91.81 $\pm$ 0.67 & 95.05 $\pm$ 0.18 & 94.49 $\pm$ 0.26 & \underline{95.12 $\pm$ 0.01} & 76.25 $\pm$ 2.83 & 94.77 $\pm$ 0.18 & \textbf{95.67 $\pm$ 0.05} \\ 
				& Flights    & 95.95 $\pm$ 0.62 & 98.22 $\pm$ 0.13 & 98.45 $\pm$ 0.01 & 91.13 $\pm$ 0.01 & 98.11 $\pm$ 0.01 & \textbf{98.93 $\pm$ 0.01} & 98.84 $\pm$ 0.01 & Timeout & 98.80$\pm$ 0.01 & \underline{98.90 $\pm$ 0.01} \\ 
				& UN Trade   & 67.44 $\pm$ 0.83 & 69.10 $\pm$ 1.67 & 68.54 $\pm$ 0.18 & 65.52 $\pm$ 0.51 & 50.00 $\pm$ 0.01 & \textbf{70.20 $\pm$ 1.44} & 50.00 $\pm$ 0.01 & 50.00 $\pm$ 0.01 & 68.24$\pm$ 0.64 & \underline{69.96 $\pm$ 1.04} \\ 
				& Contact    & 96.48 $\pm$ 0.14 & 97.54 $\pm$ 0.35 & 89.99 $\pm$ 0.34 & 93.94 $\pm$ 0.02 & 98.39 $\pm$ 0.01 & \textbf{98.53 $\pm$ 0.01} & 98.12 $\pm$ 0.02 & Timeout & 98.01$\pm$ 0.01 & \underline{98.49 $\pm$ 0.01} \\  
				\cline{2-12} 
				& Avg. Rank \rule{0pt}{8pt} &8.40&5.90&6.70&8.60&4.00&3.20&4.80&6.80&2.80&\textbf{1.40}\\ \bottomrule
	\end{tabular}}}
	\label{tab:Results_1}
\end{table*}

	\begin{table*}[h] 
		\centering
		\caption{Results for inductive dynamic link prediction with Average Precision and Area Under the ROC Curve under random negative sampling strategy.}
		%\footnotesize  % 符合AAAI 9号字体要求
		%\setlength{\tabcolsep}{2pt}  % 压缩列间距
		\resizebox{1.0\textwidth}{!}{
			\setlength{\tabcolsep}{0.9mm}{
				\begin{tabular}{c|c|cccccccccc}
					\toprule
					Metr.   \rule{0pt}{9pt}                  & Datasets    & DyRep        & TGN          & CAWN         & GraphMixer  &FreeDyG  & DyGFormer &CorDGT  & CTAN & DyGMamba & OURS   \\ \hline
					\multirow{10}{*}{AP}  
					& Wikipedia   & 92.43 $\pm$ 0.37 & 97.83 $\pm$ 0.04 & {98.24 $\pm$ 0.03} & 96.65 $\pm$ 0.02 &\underline{98.91$\pm$0.01} &98.59 $\pm$ 0.03&98.48$\pm$0.02 & 93.58 $\pm$ 0.65 & 98.77 $\pm$ 0.03 &\textbf{98.94$\pm$0.01} \\
					& Reddit      & 96.09 $\pm$ 0.11 & 97.50 $\pm$ 0.07 & {98.62 $\pm$ 0.01} & 95.26 $\pm$ 0.02 &98.84$\pm$0.02& 98.84 $\pm$ 0.02&98.82$\pm$0.01 & 80.07 $\pm$ 2.53 & \underline{98.97 $\pm$ 0.01} &\textbf{98.98$\pm$0.01} \\ 
					& MOOC        & 81.07 $\pm$ 0.44 & \underline{89.04 $\pm$ 1.17} & 81.42 $\pm$ 0.24 & 81.41 $\pm$ 0.21 &88.05$\pm$0.49&{86.96 $\pm$ 0.43}&83.74$\pm$ 0.51 & 64.99 $\pm$ 2.24 & 88.64 $\pm$ 0.08 &\textbf{90.26$\pm$0.36}\\ 
					& LastFM      & 83.02 $\pm$ 1.48 & 81.45 $\pm$ 4.29 & {89.42 $\pm$ 0.07} & 82.11 $\pm$ 0.42 &93.01 $\pm$0.23& 94.23 $\pm$ 0.09&93.03$\pm$0.18 & 60.40 $\pm$ 3.01 & \underline{94.42 $\pm$ 0.21} &\textbf{95.49$\pm$0.02} \\ 
					& Enron       & 74.55 $\pm$ 3.95 & 77.94 $\pm$ 1.02 & {86.35 $\pm$ 0.51} & 75.88 $\pm$ 0.48 &89.69 $\pm$ 0.17& \underline{89.76 $\pm$ 0.34}&\textbf{91.65$\pm$0.13} & 74.61 $\pm$ 1.64 & 89.67 $\pm$ 0.27 &88.25$\pm$0.27 \\
					& Social Evo. & 90.04 $\pm$ 0.47 & 90.77 $\pm$ 0.86 & 79.94 $\pm$ 0.18 & 91.86 $\pm$ 0.06 &93.10 $\pm$0.02& \underline{93.14 $\pm$ 0.04}&92.44$\pm$0.08 & Timeout & 93.13 $\pm$ 0.05 &\textbf{93.21$\pm$0.02} \\ 
					& UCI         & 57.48 $\pm$ 1.87 & 88.12 $\pm$ 2.05 & 92.73 $\pm$ 0.06 & 91.19 $\pm$ 0.42 &94.50$\pm$0.14& {94.54 $\pm$ 0.12}&94.70$\pm$0.10 & 49.78 $\pm$ 5.02 & \underline{94.76 $\pm$ 0.19} &\textbf{94.97$\pm$0.06} \\ 
					& Flights     & 92.88 $\pm$ 0.73 & 95.03 $\pm$ 0.60 & {97.06 $\pm$ 0.02} & 83.03 $\pm$ 0.05 &96.55$\pm$0.02& \textbf{97.79 $\pm$ 0.02}&97.46$\pm$0.04 & Timeout & 97.24$\pm$ 0.03 &\underline{97.71$\pm$0.02} \\ 
					& UN Trade    & 57.02 $\pm$ 0.69 & 58.31 $\pm$ 3.15 & \underline{65.24 $\pm$ 0.21} & 62.17 $\pm$ 0.31 &50.00$\pm$0.01  &{64.55 $\pm$ 0.62}&50.00$\pm$0.01 & 50.00 $\pm$ 0.01 & 64.25$\pm$ 0.70 &\textbf{65.40$\pm$0.59} \\ 
					& Contact     & 92.18 $\pm$ 0.41 & 93.82 $\pm$ 0.99 & 89.55 $\pm$ 0.30 & 90.59 $\pm$ 0.05 &97.66$\pm$0.02& \textbf{98.03 $\pm$ 0.02}&97.51$\pm$0.03 & Timeout & 97.42$\pm$ 0.02 &\underline{97.79$\pm$0.02} \\  
					\cline{2-12} 
					& Avg. Rank \rule{0pt}{8pt}    & 8.30&6.60&6.20&7.60&4.30&2.90&4.50&8.00&2.50&\textbf{1.70}\\ \hline
					\multirow{10}{*}{AUC}  
					& Wikipedia   & 91.49 $\pm$ 0.45 & 97.72 $\pm$ 0.03 & {98.03 $\pm$ 0.04} & 96.30 $\pm$ 0.04 &\underline{98.83$\pm$0.01} &{98.48 $\pm$ 0.03}&98.26$\pm$0.03 & 92.59 $\pm$ 0.70 & 98.72 $\pm$ 0.03 &\textbf{98.93$\pm$0.01} \\ 
					& Reddit      & 96.05 $\pm$ 0.12 & 97.39 $\pm$ 0.07 & {98.42 $\pm$ 0.02} & 94.97 $\pm$ 0.05 &98.73$\pm$0.02 &{98.71 $\pm$ 0.01}&98.67$\pm$0.08 & 82.35 $\pm$ 4.03 & \underline{98.88 $\pm$ 0.01} &\textbf{98.89$\pm$0.01}\\ 
					& MOOC        & 84.03 $\pm$ 0.49 & \underline{91.24 $\pm$ 0.99} & 81.86 $\pm$ 0.25 & 82.77 $\pm$ 0.24 &88.30$\pm$0.31 &{87.62 $\pm$ 0.51}&84.90$\pm$ 0.55 & 66.38 $\pm$ 1.59 & 89.34 $\pm$ 0.12 &\textbf{91.81$\pm$0.25} \\ 
					& LastFM      & 82.24 $\pm$ 1.51 & 82.61 $\pm$ 3.15 & {87.82 $\pm$ 0.12} & 80.37 $\pm$ 0.18 &92.11 $\pm$0.02& 94.08 $\pm$ 0.08&93.01$\pm$0.18 & 61.49 $\pm$ 2.78 & \underline{94.37 $\pm$ 0.13} &\textbf{95.31$\pm$0.02}\\ 
					& Enron       & 76.34 $\pm$ 4.20 & 78.83 $\pm$ 1.11 & {87.02 $\pm$ 0.50} & 76.51 $\pm$ 0.71 &89.51 $\pm$ 0.20& \underline{90.69 $\pm$ 0.26}&\textbf{92.64$\pm$0.19} & 75.23 $\pm$ 2.24 & 89.76 $\pm$ 0.21 &89.56$\pm$0.13 \\ 
					& Social Evo. & 91.18 $\pm$ 0.49 & 93.43 $\pm$ 0.59 & 84.73 $\pm$ 0.27 & {94.09 $\pm$ 0.07} &95.36 $\pm$0.10 & {95.29 $\pm$ 0.03}&\underline{95.69$\pm$0.10} & Timeout & 95.36 $\pm$ 0.04 &\textbf{95.70$\pm$0.02} \\ 
					& UCI         & 58.08 $\pm$ 1.81 & 86.68 $\pm$ 2.29 & {90.40 $\pm$ 0.11} & 89.30 $\pm$ 0.57 &93.19$\pm$0.10& {92.63 $\pm$ 0.13}&\underline{93.20$\pm$0.09} & 48.58 $\pm$ 6.02 & 92.70 $\pm$ 0.19 &\textbf{93.58$\pm$0.11} \\ 
					& Flights     & 93.56 $\pm$ 0.70 & 95.92 $\pm$ 0.43 & {96.86 $\pm$ 0.02} & 82.27 $\pm$ 0.06 &96.14$\pm$0.02& \underline{97.80 $\pm$ 0.02}&\textbf{98.12$\pm$0.04} & Timeout & 97.29$\pm$ 0.02 &97.73$\pm$0.02 \\ 
					& UN Trade    & 58.82 $\pm$ 0.98 & 59.99 $\pm$ 3.50 & 67.05 $\pm$ 0.21 & 63.48 $\pm$ 0.37 &50.00$\pm$0.01 &{\underline{67.25 $\pm$ 1.05}}&50.00$\pm$0.01 & 50.00 $\pm$ 0.01 & 66.76$\pm$ 0.61 &\textbf{67.49$\pm$1.14} \\ 
					& Contact     & 91.89 $\pm$ 0.38 & 94.84 $\pm$ 0.75 & 89.07 $\pm$ 0.34 & 92.83 $\pm$ 0.05 &98.12$\pm$0.01& \underline{98.30 $\pm$ 0.02}&97.86 $\pm$ 0.02 & Timeout & 98.18$\pm$ 0.01 &\textbf{98.32$\pm$0.02} \\  
					\cline{2-12} 
					& Avg. Rank \rule{0pt}{8pt}    & 8.20&6.40&6.50&7.60&4.30&3.40&4.00&8.00&2.60&\textbf{1.60} \\ \bottomrule
		\end{tabular}}}
		\label{tab:Results_2}
	\end{table*}
\subsection{Performance Comparison Analysis}
\begin{figure}[h]
	\centering
	\includegraphics[width=0.4\textwidth]{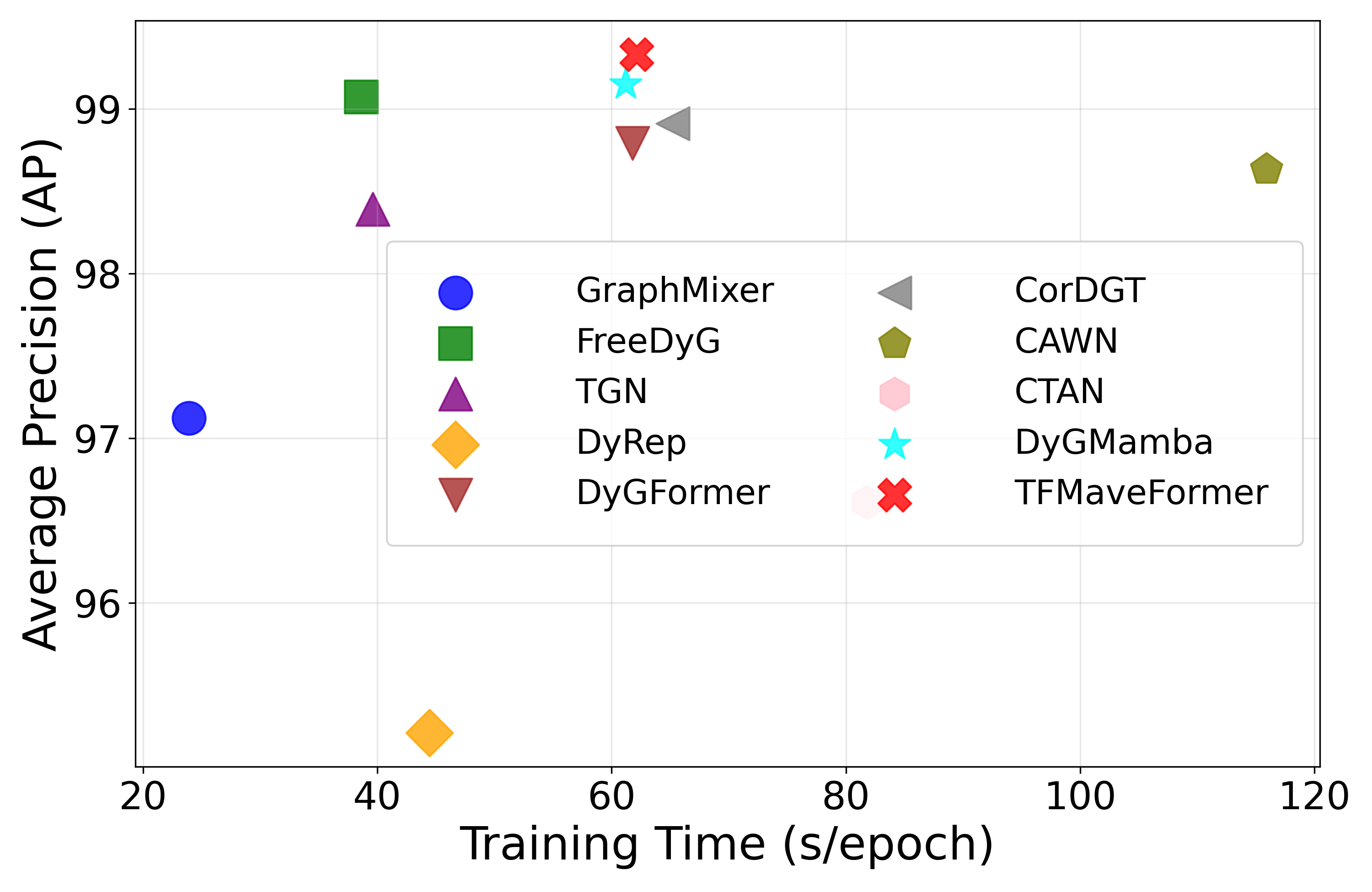}
	\includegraphics[width=0.4\textwidth]{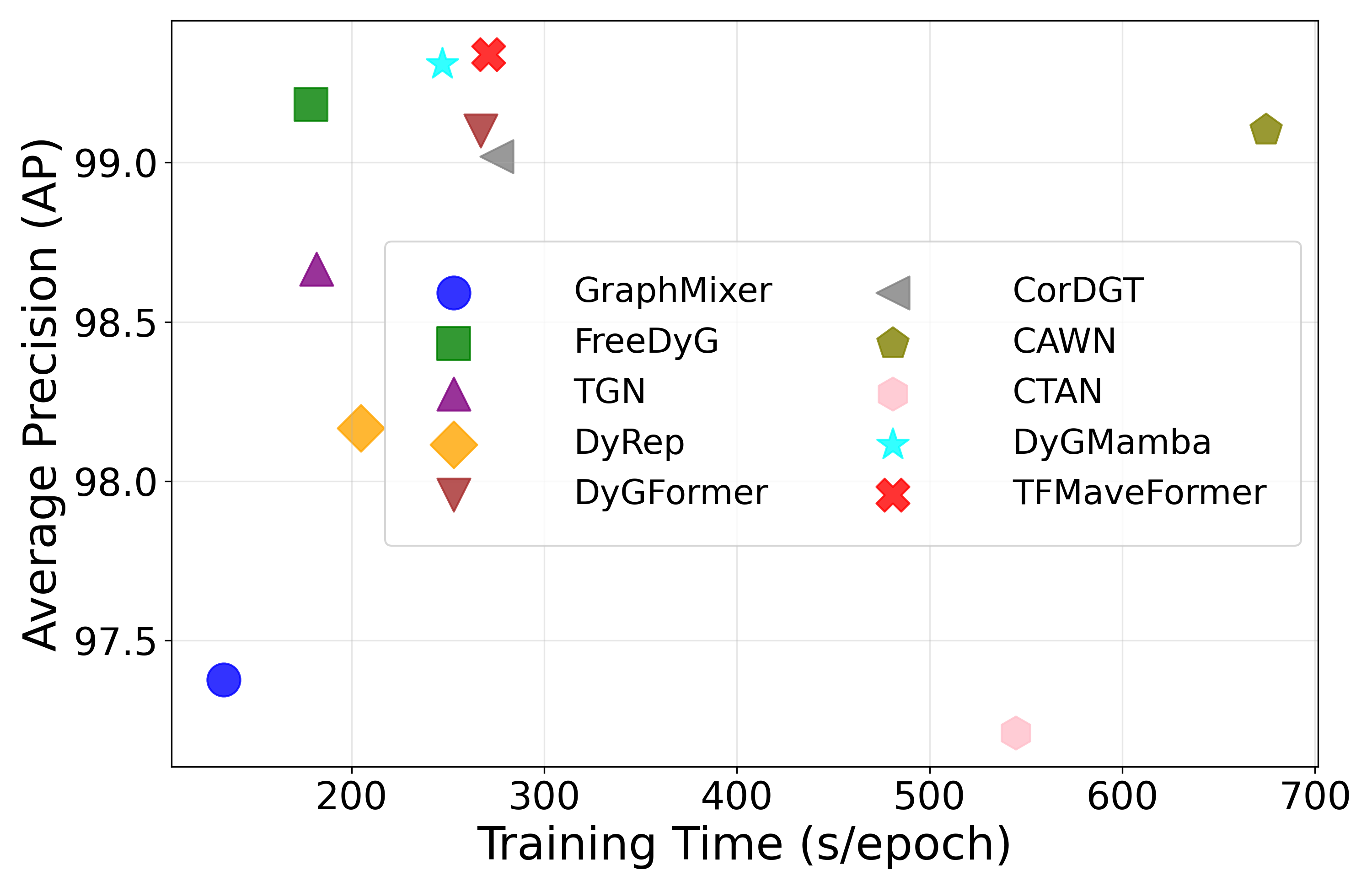}
	\caption{Comparison performance and training time per epoch on Wikipedia and Reddit.(a) Wikipedia, (b) Reddit.}
	\label{fig:performance_time}
\end{figure}

Figure~\ref{fig:performance_time} demonstrates the performance-efficiency trade-off across different dynamic graph learning methods on Wikipedia and Reddit datasets. While recent approaches like FreeDyG, DyGFormer, and CorDGT, CTAN, DyMamba achieve superior predictive accuracy compared to earlier methods such as DyRep and TGAT, they require substantially longer training time per epoch. Our TFWaveFormer achieves the optimal balance, delivering the best performance with training efficiency comparable to state-of-the-art methods.

Tables~\ref{tab:Results_1} and~\ref{tab:Results_2} present comprehensive experimental results across 10 diverse benchmark datasets with random negative sampling under both transductive and inductive settings, respectively. TFWaveFormer achieves consistently superior results with average rankings of 1.20 and 1.40 for AP and AUC metrics in transductive settings, and 1.70 and 1.60 in inductive settings. This represents substantial improvements over the second-best performing methods, with DyGFormer achieving rankings of 3.10/2.50 in transductive and 2.40/2.80 in inductive settings respectively.
Our method demonstrates remarkable consistency across diverse graph types and scales. On social networks including Wikipedia and Reddit, TFWaveFormer achieves 99.33\% and 99.32\% AP scores respectively, establishing new performance benchmarks. On temporal interaction graphs such as MOOC and LastFM, it shows significant improvements of 2.01\% and 1.29\% AP over the second-best methods. Even on challenging sparse datasets like UN Trade, our approach maintains competitive performance with 66.06\% AP compared to 65.39\% from the previous best method.
The inductive setting results validate TFWaveFormer's superior generalization ability to unseen nodes and temporal patterns. Notable improvements include 1.17\% AP gain on LastFM and consistent top-2 performance across all datasets. The consistently low standard deviations, typically below 0.05 across multiple runs, indicate stable and reliable performance, with our method showing the most consistent results among all compared approaches.

This superior performance stems from TFWaveFormer's dual architectural innovations. The multi-resolution wavelet decomposition captures temporal patterns across multiple scales simultaneously, enabling comprehensive analysis of both short-term interactions and long-term evolutionary trends. The adaptive gated fusion mechanism dynamically balances temporal and frequency domain features based on input characteristics, allowing the model to emphasize the most relevant information for different types of dynamic graph behaviors. These innovations collectively enable more effective modeling of complex temporal dependencies in dynamic graphs, resulting in enhanced predictive accuracy and robust generalization capabilities.
	
\subsection{Ablation analysis} 
	\begin{figure*}[t]
		\centering
		\begin{subfigure}{0.23\textwidth}
			\includegraphics[width=\textwidth]{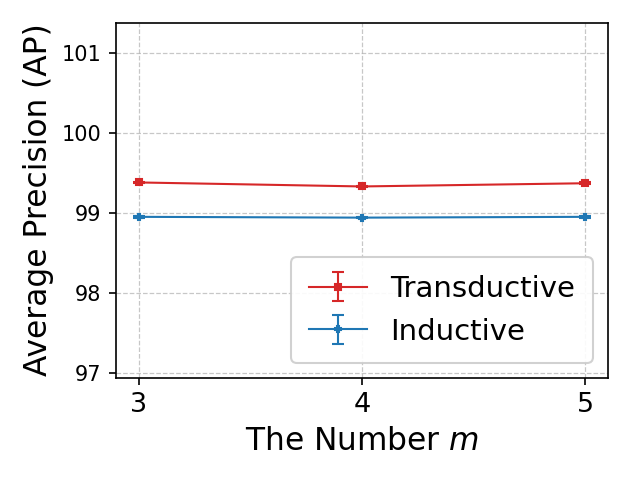}
		%	\caption{}
		\end{subfigure}
		\begin{subfigure}{0.23\textwidth}
			\includegraphics[width=\textwidth]{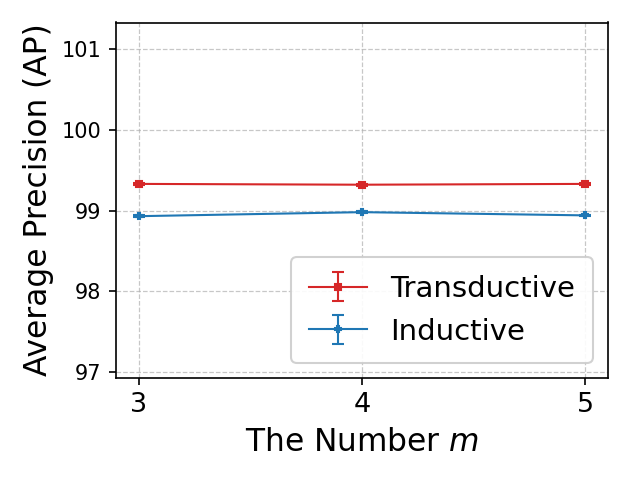}
		%	\caption{}
		\end{subfigure}
		\begin{subfigure}{0.23\textwidth}
			\includegraphics[width=\textwidth]{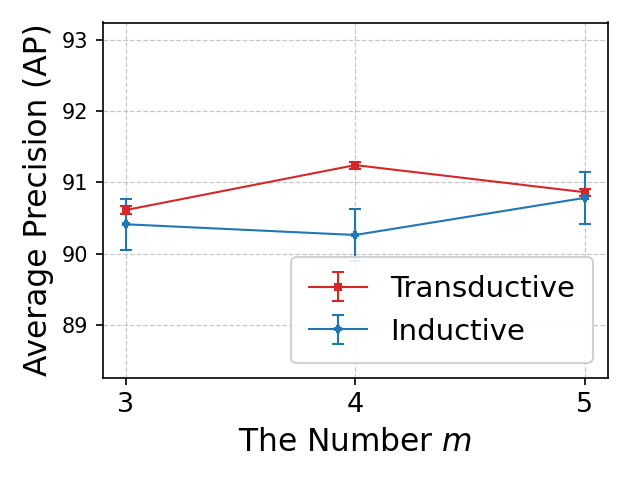}
		%	\caption{}
		\end{subfigure}
		\begin{subfigure}{0.23\textwidth}
			\includegraphics[width=\textwidth]{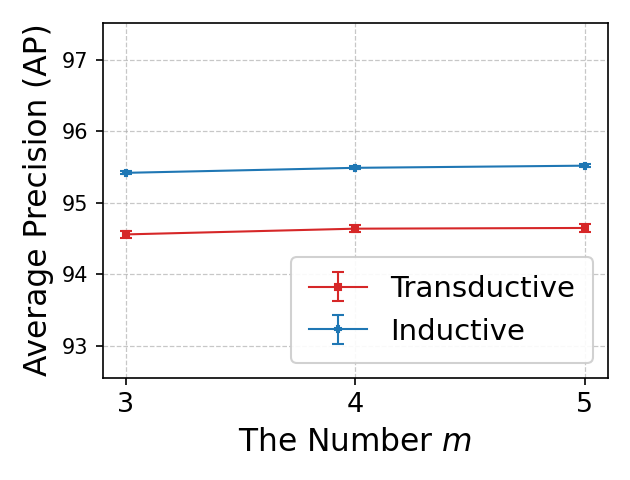}
		%	\caption{}
		\end{subfigure}
		\begin{subfigure}{0.23\textwidth}
			\includegraphics[width=\textwidth]{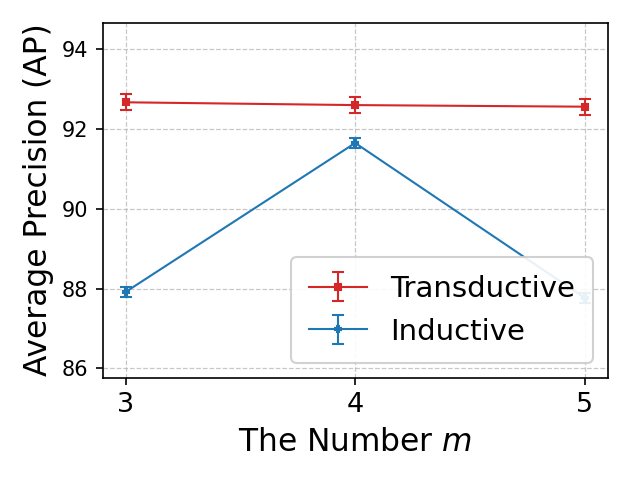}
		%	\caption{}
		\end{subfigure}
		\begin{subfigure}{0.23\textwidth}
			\includegraphics[width=\textwidth]{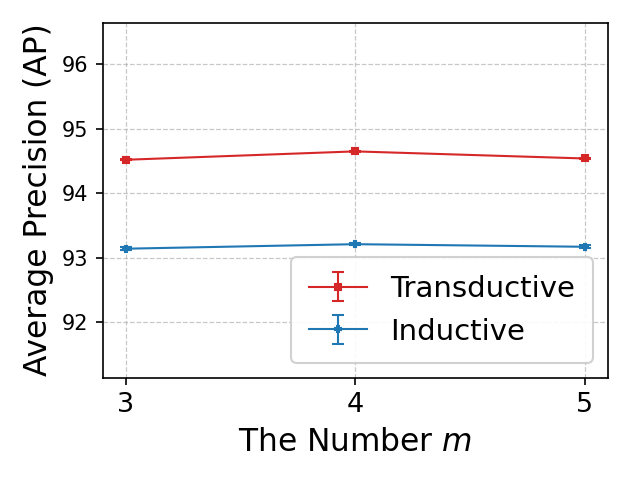}
		%	\caption{}
		\end{subfigure}
		\begin{subfigure}{0.23\textwidth}
			\includegraphics[width=\textwidth]{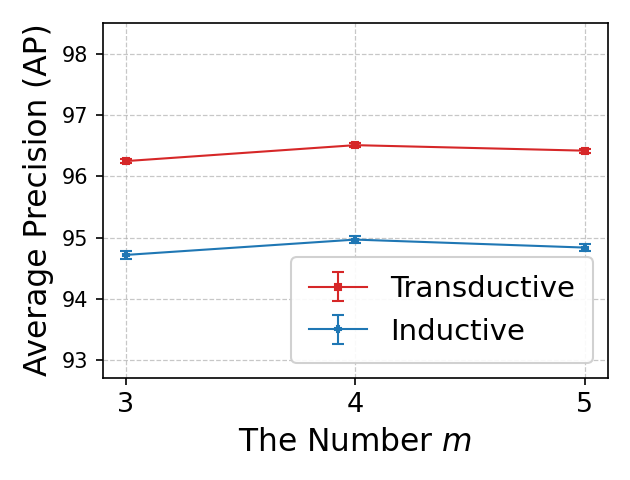}
		%	\caption{}
		\end{subfigure}
		\begin{subfigure}{0.23\textwidth}
			\includegraphics[width=\textwidth]{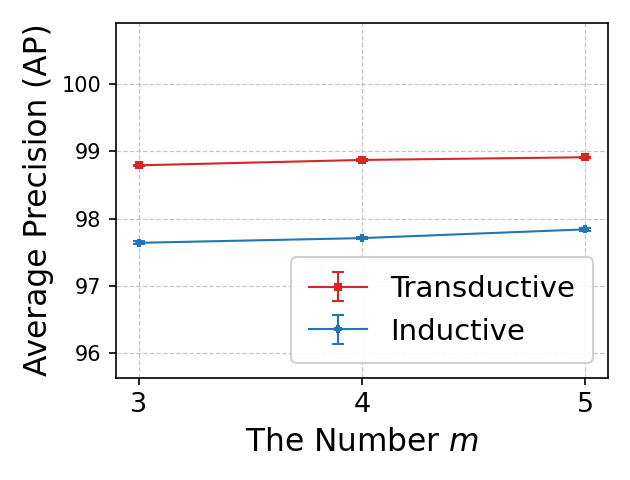}
%			\caption{}
		\end{subfigure}
		\caption{Results of hyper-parameters sensitivity to the number of wavelet convolution kernels $m$ in dynamic link prediction experiments across different datasets. (a) Wikipedia, (b) Reddit, (c) MOOC, (d) LastFM, (e) Enron, (f) Social Evolution, (g) UCI, (h) Flights.}
		\label{fig:wavelet_kernel1}
	\end{figure*}
	\begin{figure}[h]
		\centering
		\includegraphics[width=0.4\textwidth]{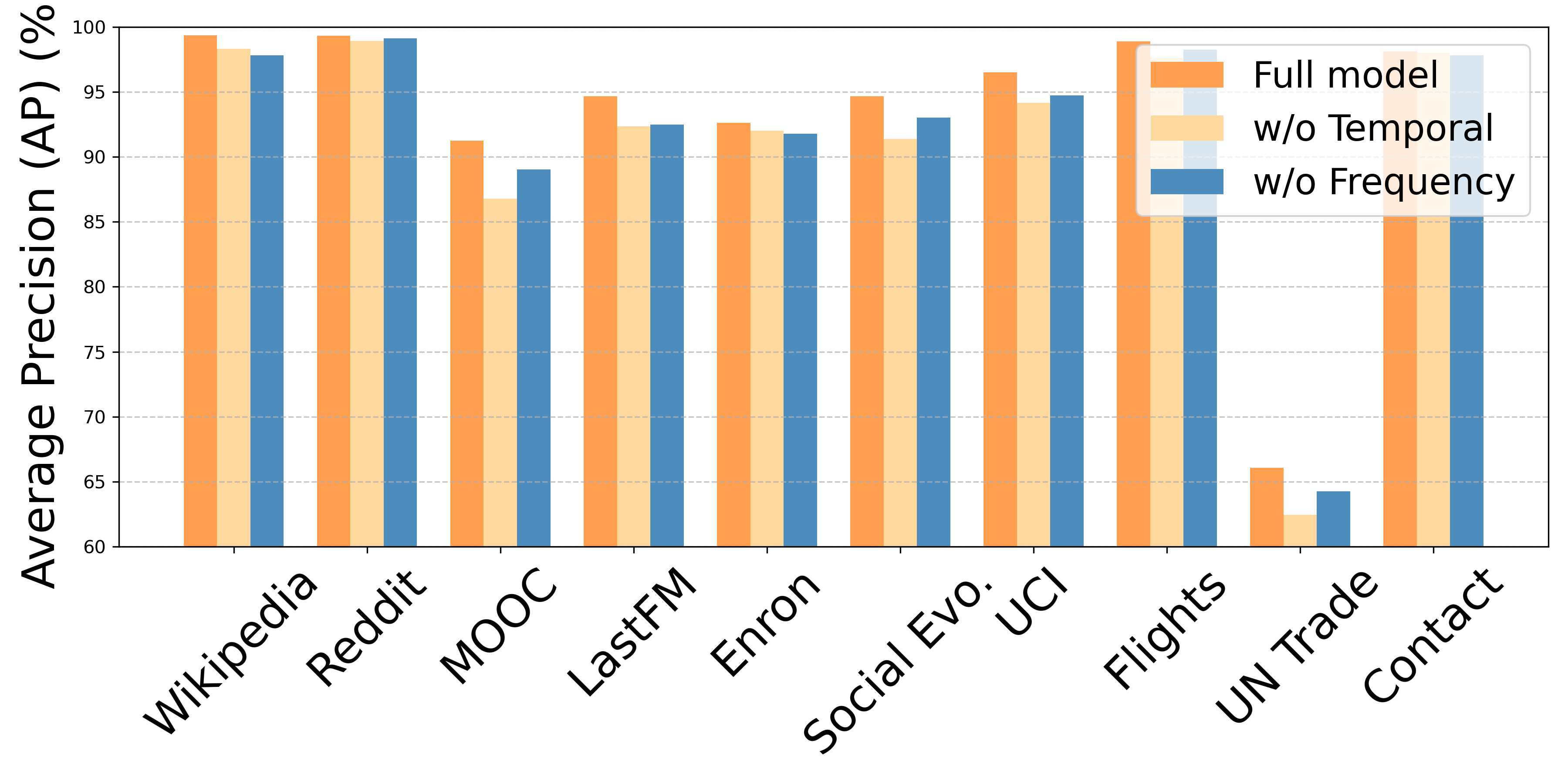}
		\includegraphics[width=0.4\textwidth]{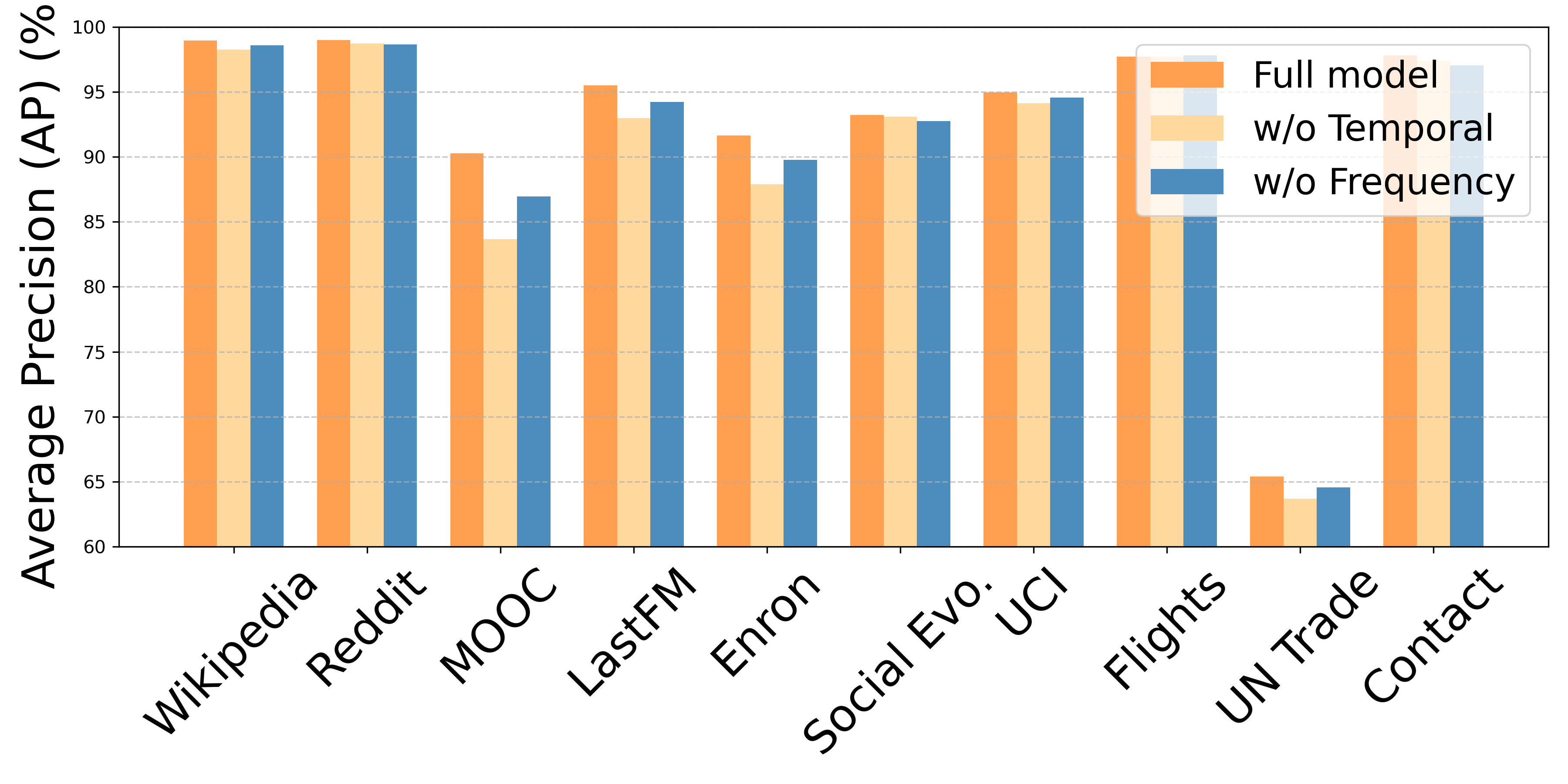}
		\caption{Results on ablation study for dynamic link prediction Under differrent settings. (a) Transductive setting, (b) Inductive setting.}
		\label{fig:ablation}
	\end{figure}
	%In Fig.~\ref{fig:ablation}, we present a detailed ablation study of TFWaveFormer under both transductive and inductive settings to analyze the contribution of each core component. Specifically, the variant "w/o Temporal" removes the temporal modeling branch along with the NIF encoder, effectively discarding explicit temporal context. In contrast, the variant "w/o Frequency" eliminates the multi-level wavelet transform module, directly feeding the features into the Transformer without frequency decomposition. Among these variants, "w/o Frequency" causes a notable performance degradation, underscoring the effectiveness and necessity of the proposed wavelet-based frequency modeling. Similarly, "w/o Temporal" also leads to consistent performance drops across most datasets, which highlights the importance of preserving raw temporal signals and their synergy with frequency representations in capturing fine-grained dynamics and non-periodic interactions in evolving graphs.
	Figure~\ref{fig:ablation} presents a comprehensive ablation study evaluating the contribution of each core component in TFWaveFormer across 10 benchmark datasets under both transductive and inductive settings. We analyze two critical variants: "w/o Temporal" removes the temporal modeling branch including the NIF encoder, while "w/o Frequency" discards the multi-level wavelet transform module and directly feeds features into the Transformer without frequency decomposition.
	
	The results demonstrate that both components are essential for optimal performance, with frequency domain modeling showing particularly critical importance. The "w/o Frequency" variant exhibits substantial performance degradation across all datasets, especially on challenging datasets such as MOOC and LastFM, highlighting the necessity of multi-scale temporal pattern capture through wavelet decomposition.
	The temporal component also proves indispensable, with "w/o Temporal" showing consistent performance deterioration across most datasets. While the performance impact is relatively modest on interaction-rich datasets like Wikipedia and Reddit due to abundant structural information, the degradation becomes more pronounced on sparse temporal graphs such as UN Trade and Contact, emphasizing the importance of preserving raw temporal signals for capturing fine-grained dynamics.
	
	The ablation results reveal dataset-dependent sensitivity to different components. Dense social networks show greater resilience to temporal component removal, while sparse interaction networks demonstrate higher dependency on both temporal and frequency modeling. This validates our design philosophy that effective dynamic graph learning requires synergistic integration of temporal-domain and frequency-domain representations to handle diverse graph characteristics and interaction patterns.
	
	\subsection{Robustness analysis}
	%The task of dynamic link prediction encompasses three progressively challenging negative sampling strategies: random (rnd), historical (hist), and inductive (ind). Tables~\ref{tab:Results_1} and~\ref{tab:Results_2} report results under the basic rnd setting, while Table~\ref{tab:experimental_results} presents performance under the more difficult hist and ind scenarios. Although we provide a concise overview due to space limitations, it is evident that our method consistently achieves strong results, even under the more complex hist and ind settings. This demonstrates the proposed model’s excellent generalization ability and robustness when facing unseen links or dynamic structural shifts in temporal networks. More details of the results can be found in Appendix.B.
	Dynamic link prediction encompasses three progressively challenging negative sampling strategies that reflect different real-world scenarios: random, history, inductive. Beyond the random negative sampling strategy reported in Tables~\ref{tab:Results_1} and~\ref{tab:Results_2}, we evaluate TFWaveFormer under more rigorous conditions using historical and inductive negative sampling approaches. Table~\ref{tab:experimental_results} presents comprehensive results across all 10 benchmark datasets under these challenging settings.
	The historical negative sampling strategy (hist) selects negative samples from nodes that have previously interacted but not at the current timestamp, creating more realistic and difficult prediction scenarios. The inductive negative sampling strategy (ind) further increases complexity by selecting negatives from completely unseen node pairs during training. These strategies better reflect real-world applications where models must distinguish between likely future interactions and plausible but incorrect predictions based on historical patterns.
	
	Results demonstrate that TFWaveFormer maintains robust performance across all challenging scenarios. Under the historical strategy, our method achieves strong AP scores ranging from 60.16\% on sparse datasets like UN Trade to 97.85\% on dense interaction networks like Contact. The inductive strategy, while generally more challenging, shows consistent performance with scores spanning from 61.37\% to 97.19\% across different dataset characteristics. Notably, some datasets such as Reddit and Social Evolution even show improved performance under inductive sampling, indicating our model's ability to effectively generalize beyond training distributions.
	The consistent performance across different negative sampling strategies validates the robustness of our dual-domain modeling approach. The multi-resolution wavelet decomposition enables effective capture of complex temporal patterns regardless of sampling difficulty, while the adaptive fusion mechanism ensures optimal feature integration across diverse prediction scenarios. This demonstrates TFWaveFormer's practical applicability in real-world dynamic graph applications where negative sampling strategies may vary significantly. Due to space limitations, we only provide a brief overview; more detailed results are provided in Appendix A.2.
	%Table~\ref{tab:experimental_results} incorporating three increasingly challenging negative sampling strategies: random (rnd), historical (hist), and inductive (ind). For node classification, considering class imbalance, we adopt AUC-ROC as the evaluation metric.
	
	% We further investigate the sensitivity of TFWaveFormer to its key hyperparameters and summarize the following insights. As shown in Fig.~\ref{fig:wavelet_kernel}, for each dataset, there exists a performance-optimal value for the number of wavelet convolutional kernels $m$ used in wavelet decomposition. Different datasets exhibit varying optimal values due to their intrinsic distributions of nodes and temporal link patterns. For example, Wikipedia and Reddit, as large-scale social networks with dense interactions and periodic activities, achieve optimal performance at $m=5$, which captures both short-term bursts and long-term cycles. In contrast, Enron and UCI, representing email and messaging networks with sparser and more irregular activity, reach their best performance at $m=3$, as fewer scales suffice to model their simpler dynamics. Other datasets demonstrate dataset-specific optimal points. Due to time constraints, we only tested values of $m$ up to 5. It is possible that more complex datasets may benefit from higher values, which we leave for future exploration.
	
	%incorporating three increasingly challenging negative sampling strategies: random (rnd), historical (hist), and inductive (ind). For node classification, considering class imbalance, we adopt AUC-ROC as the evaluation metric.
	\begin{table}[h]
		\centering
		\caption{Results for dynamic link prediction with AP under historical and inductive negative sampling strategies.}
		\begin{tabular}{lcccc}
			\toprule
			\multirow{2}{*}{Datasets}& \multicolumn{2}{c}{transductive} & \multicolumn{2}{c}{inductive} \\
			\cmidrule(lr){2-3} \cmidrule(lr){4-5}
			& hist & ind & hist & ind \\
			\midrule
			Wikipedia & 86.21 & 73.36 & 70.08 & 70.07 \\
			Reddit & 83.77 & 90.88 & 68.50 & 68.54 \\
			MOOC & 88.31 & 82.46 & 79.97 & 79.99 \\
			LastFM & 84.80 & 75.52 & 79.52 & 79.52 \\
			Enron & 78.64 & 84.23 & 81.26 & 81.26 \\
			Social Evo. & 96.96 & 97.19 & 95.85 & 95.86 \\
			UCI & 90.19 & 86.95 & 87.57 & 87.59 \\
			Flights & 65.31 & 71.20 & 56.68 & 56.72 \\
			UN Trade & 60.16 & 61.37 & 53.88 & 53.90 \\
			Contact & 97.85 & 96.10 & 95.27 & 94.69 \\
			\bottomrule
		\end{tabular}
		\label{tab:experimental_results}
	\end{table}
	\subsection{Parameter analysis} 
	We further investigate the sensitivity of TFWaveFormer to its key hyper-parameters and summarize the following insights. As shown in Fig.~\ref{fig:wavelet_kernel1}, for each dataset, there exists a performance-optimal value for the number of wavelet convolutional kernels $m$ used in wavelet decomposition. Different datasets exhibit varying optimal values due to their intrinsic distributions of nodes and temporal link patterns. 
	For example, Wikipedia and Reddit, as large-scale social networks with dense interactions and periodic activities, achieve optimal performance at $m=5$, which captures both short-term bursts and long-term cycles. In contrast, Enron and UCI, representing email and messaging networks with sparser and more irregular activity, reach their best performance at $m=3$, as fewer scales suffice to model their simpler dynamics. 
	We observe that performance degradation is more pronounced when $m$ deviates from the optimal value for complex datasets compared to sparse datasets, indicating higher sensitivity to multi-scale modeling in temporally complex networks.
	Other datasets demonstrate dataset-specific optimal points. 
	Notably, we find a correlation between dataset temporal complexity and optimal $m$ values: datasets with more diverse temporal patterns require higher $m$ values for optimal performance. 
	Due to time constraints, we only tested values of $m$ up to 5. It is possible that more complex datasets may benefit from higher values, which we leave for future exploration. Due to space constraints, we present a subset of the results here, with the remaining results available in Appendix A.3.
	
	\section{Conclusion}
	%In this paper, we proposed TFWaveFormer, a novel framework that jointly models temporal and spectral representations for dynamic link prediction. 
	This work proposes a temporal-frequency collaborative framework TFWaveFormer designed to overcome key limitations in dynamic link prediction. By integrating micro-level temporal dynamics with macro-level evolutionary trends via multi-scale modeling, the framework captures both transient events and periodic patterns in evolving graphs. Extensive experiments on ten real-world datasets show that TFWaveFormer consistently achieves state-of-the-art results across both transductive and inductive settings. Furthermore, our analysis highlights that the optimal wavelet scales vary by dataset, underscoring the need for adaptive configuration based on graph characteristics.
	%In this paper, we address the critical limitations of existing dynamic link prediction methods by proposing TFWaveFormer, a temporal-frequency collaborative framework that unifies micro-level temporal dynamics with macro-level evolutionary trends through multi-scale modeling. By introducing a multi-level wavelet transform module and a gated fusion mechanism, our approach effectively captures both transient events and periodic patterns in evolving graphs. Extensive experiments on 10 real-world datasets demonstrate that TFWaveFormer achieves state-of-the-art performance, ranking best on average across both transductive and inductive settings. Notably, our findings also reveal that optimal wavelet scales vary across datasets, suggesting the importance of adaptive configuration based on graph characteristics.	
%	\bibliography{aaai2026}
%	%	\bibliography{iclr2024_conference}
%	\bibliographystyle{aaai26}
\section*{ACKNOWLEDGMENTS}
This research was supported by the National Natural Science Foundation of China (Grants No. 62233004, 62273090, and T2541017), the Youth Scientist Project of the Ministry of Science and Technology of China (Grant No. 2025YFF0524100), the Zhishan Youth Scholar Program of the Southeast University, the Jiangsu Provincial Scientific Research Center of Applied Mathematics (Grant No. BK20233002), the Basic Research Program of Jiangsu (Grant No. BK20253018), the Open Research Project of the State Key Laboratory of Industrial Control Technology, China (Grant No. ICT2025B54).
%This research was supported by the National Science and Technology Major Project of the Ministry of Science and Technology of China (Grant No. 2024ZD0608104), the Youth Scientist Project of the Ministry of Science and Technology of China (Grant No. 2025YFF0524100), the National Natural Science Foundation of China (Grants No. 62233004, 62273090, 62073076, and T2541017).
%%, the Zhishan Youth Scholar Program of the Southeast University, the Jiangsu Provincial Scientific Research Center of Applied Mathematics (Grant No. BK20233002), the Natural Science Foundation of Jiangsu Province of China (Grant No. BK20253020), the Open Research Project of the State Key Laboratory of Industrial Control Technology, China (Grant No. ICT2025B54).
\bibliographystyle{unsrt}
\bibliography{biblist}

\begin{thebibliography}{10}

\bibitem{DBLP:journals/jmlr/KazemiGJKSFP20}
Seyed~Mehran Kazemi, Rishab Goel, Kshitij Jain, Ivan Kobyzev, Akshay Sethi, Peter Forsyth, and Pascal Poupart.
\newblock Representation learning for dynamic graphs: {A} survey.
\newblock {\em J. Mach. Learn. Res.}, 21:70:1--70:73, 2020.

\bibitem{ekle2024anomaly}
Ocheme~Anthony Ekle and William Eberle.
\newblock Anomaly detection in dynamic graphs: A comprehensive survey.
\newblock {\em ACM Transactions on Knowledge Discovery from Data}, 18(8):1--44, 2024.

\bibitem{alvarez2021evolutionary}
Unai Alvarez-Rodriguez, Federico Battiston, Guilherme~Ferraz de~Arruda, Yamir Moreno, Matja{\v{z}} Perc, and Vito Latora.
\newblock Evolutionary dynamics of higher-order interactions in social networks.
\newblock {\em Nature Human Behaviour}, 5(5):586--595, 2021.

\bibitem{DrugInteraction}
E.~Amiri Souri, Roman Laddach, S.~N. Karagiannis, Lazaros~G. Papageorgiou, and Sophia Tsoka.
\newblock Novel drug-target interactions via link prediction and network embedding.
\newblock {\em {BMC} Bioinform.}, 23(1):121, 2022.

\bibitem{Zhang2020}
Muhan Zhang and Yixin Chen.
\newblock Inductive matrix completion based on graph neural networks.
\newblock In {\em ICLR}, 2020.

\bibitem{NBFNet}
Zhaocheng Zhu, Zuobai Zhang, Louis{-}Pascal A.~C. Xhonneux, and Jian Tang.
\newblock Neural bellman-ford networks: {A} general graph neural network framework for link prediction.
\newblock In {\em NeurIPS}, pages 29476--29490, 2021.

\bibitem{bai2024dynamic}
Jiaru Bai, Sebastian Mosbach, Connor~J Taylor, Dogancan Karan, Kok~Foong Lee, Simon~D Rihm, Jethro Akroyd, Alexei~A Lapkin, and Markus Kraft.
\newblock A dynamic knowledge graph approach to distributed self-driving laboratories.
\newblock {\em Nature Communications}, 15(1):462, 2024.

\bibitem{zhang2024llm4dyg}
Zeyang Zhang, Xin Wang, Ziwei Zhang, Haoyang Li, Yijian Qin, and Wenwu Zhu.
\newblock Llm4dyg: can large language models solve spatial-temporal problems on dynamic graphs?
\newblock In {\em Proceedings of the 30th ACM SIGKDD Conference on Knowledge Discovery and Data Mining}, pages 4350--4361, 2024.

\bibitem{DBLP:conf/kdd/KumarZL19}
Srijan Kumar, Xikun Zhang, and Jure Leskovec.
\newblock Predicting dynamic embedding trajectory in temporal interaction networks.
\newblock In {\em Proceedings of the 25th {ACM} {SIGKDD} International Conference on Knowledge Discovery {\&} Data Mining}, pages 1269--1278. {ACM}, 2019.

\bibitem{DBLP:journals/corr/abs-2006-10637}
Emanuele Rossi, Ben Chamberlain, Fabrizio Frasca, Davide Eynard, Federico Monti, and Michael Bronstein.
\newblock Temporal graph networks for deep learning on dynamic graphs.
\newblock In {\em ICML 2020 Workshop on Graph Representation Learning}, 2020.

\bibitem{DBLP:conf/sigir/0001GRTY20}
Yao Ma, Ziyi Guo, Zhaochun Ren, Jiliang Tang, and Dawei Yin.
\newblock Streaming graph neural networks.
\newblock In {\em Proceedings of the 43rd International {ACM} {SIGIR} conference on research and development in Information Retrieval}, pages 719--728. {ACM}, 2020.

\bibitem{weisfeiler1968reduction}
Boris Weisfeiler and Andrei Leman.
\newblock The reduction of a graph to canonical form and the algebra which appears therein.
\newblock {\em nti, Series}, 2(9):12--16, 1968.

\bibitem{aamand2022exponentially}
Anders Aamand, Justin Chen, Piotr Indyk, Shyam Narayanan, Ronitt Rubinfeld, Nicholas Schiefer, Sandeep Silwal, and Tal Wagner.
\newblock Exponentially improving the complexity of simulating the weisfeiler-lehman test with graph neural networks.
\newblock {\em Advances in Neural Information Processing Systems}, 35:27333--27346, 2022.

\bibitem{DBLP:conf/sigmod/WangLLXYWWCYSG21}
Xuhong Wang, Ding Lyu, Mengjian Li, Yang Xia, Qi~Yang, Xinwen Wang, Xinguang Wang, Ping Cui, Yupu Yang, Bowen Sun, and Zhenyu Guo.
\newblock {APAN:} asynchronous propagation attention network for real-time temporal graph embedding.
\newblock In {\em International Conference on Management of Data}, pages 2628--2638. {ACM}, 2021.

\bibitem{luo2022neighborhoodaware}
Yuhong Luo and Pan Li.
\newblock Neighborhood-aware scalable temporal network representation learning.
\newblock In {\em The First Learning on Graphs Conference}, 2022.

\bibitem{DBLP:conf/iclr/XuRKKA20}
Da~Xu, Chuanwei Ruan, Evren K{\"{o}}rpeoglu, Sushant Kumar, and Kannan Achan.
\newblock Inductive representation learning on temporal graphs.
\newblock In {\em 8th International Conference on Learning Representations}. OpenReview.net, 2020.

\bibitem{yu2023towards}
Le~Yu, Leilei Sun, Bowen Du, and Weifeng Lv.
\newblock Towards better dynamic graph learning: New architecture and unified library.
\newblock {\em Advances in Neural Information Processing Systems}, 36:67686--67700, 2023.

\bibitem{velivckovic2017graph}
Petar Veli{\v{c}}kovi{\'c}, Guillem Cucurull, Arantxa Casanova, Adriana Romero, Pietro Lio, and Yoshua Bengio.
\newblock Graph attention networks.
\newblock {\em arXiv preprint arXiv:1710.10903}, 2017.

\bibitem{wang2024large}
Jiapu Wang, Sun Kai, Linhao Luo, Wei Wei, Yongli Hu, Alan Wee-Chung Liew, Shirui Pan, and Baocai Yin.
\newblock Large language models-guided dynamic adaptation for temporal knowledge graph reasoning.
\newblock {\em Advances in Neural Information Processing Systems}, 37:8384--8410, 2024.

\bibitem{Shi_Fan_Kwok_2020}
Han Shi, Haozheng Fan, and James~T. Kwok.
\newblock Effective decoding in graph auto-encoder using triadic closure.
\newblock {\em Proceedings of the AAAI Conference on Artificial Intelligence}, 34(01):906--913, Apr. 2020.

\bibitem{DBLP:conf/nips/SouzaMKG22}
Amauri~H. Souza, Diego Mesquita, Samuel Kaski, and Vikas Garg.
\newblock Provably expressive temporal graph networks.
\newblock In {\em NeurIPS}, 2022.

\bibitem{DBLP:conf/wsdm/SankarWGZY20}
Aravind Sankar, Yanhong Wu, Liang Gou, Wei Zhang, and Hao Yang.
\newblock Dysat: Deep neural representation learning on dynamic graphs via self-attention networks.
\newblock In {\em The Thirteenth {ACM} International Conference on Web Search and Data Mining}, pages 519--527. {ACM}, 2020.

\bibitem{cong2023do}
Weilin Cong, Si~Zhang, Jian Kang, Baichuan Yuan, Hao Wu, Xin Zhou, Hanghang Tong, and Mehrdad Mahdavi.
\newblock Do we really need complicated model architectures for temporal networks?
\newblock In {\em International Conference on Learning Representations}, 2023.

\bibitem{DBLP:conf/iclr/WangCLL021}
Yanbang Wang, Yen{-}Yu Chang, Yunyu Liu, Jure Leskovec, and Pan Li.
\newblock Inductive representation learning in temporal networks via causal anonymous walks.
\newblock In {\em 9th International Conference on Learning Representations}. OpenReview.net, 2021.

\bibitem{wang2025dynamic}
Zhe Wang, Sheng Zhou, Jiawei Chen, Zhen Zhang, Binbin Hu, Yan Feng, Chun Chen, and Can Wang.
\newblock Dynamic graph transformer with correlated spatial-temporal positional encoding.
\newblock In {\em Proceedings of the Eighteenth ACM International Conference on Web Search and Data Mining}, pages 60--69, 2025.

\bibitem{xu2024revisiting}
Fan Xu, Nan Wang, Hao Wu, Xuezhi Wen, Xibin Zhao, and Hai Wan.
\newblock Revisiting graph-based fraud detection in sight of heterophily and spectrum.
\newblock In {\em Proceedings of the AAAI conference on artificial intelligence}, volume~38, pages 9214--9222, 2024.

\bibitem{10.1145/3583780.3615067}
Bin Wu, Xinyu Yao, Boyan Zhang, Kuo-Ming Chao, and Yinsheng Li.
\newblock Splitgnn: Spectral graph neural network for fraud detection against heterophily.
\newblock In {\em Proceedings of the 32nd ACM International Conference on Information and Knowledge Management}, CIKM '23, 2023.

\bibitem{xiong2025spectral}
Baisen Xiong, Sijie Wen, Ping Lu, and Kaibiao Lin.
\newblock A spectral domain graph transformer model with position-encoded information.
\newblock {\em IAENG International Journal of Computer Science}, 52(6), 2025.

\bibitem{shalby2025comprehensive}
Esraa~M Shalby, Almoataz~Y Abdelaziz, Eman~S Ahmed, and Basem Abd-Elhamed~Rashad.
\newblock A comprehensive guide to selecting suitable wavelet decomposition level and functions in discrete wavelet transform for fault detection in distribution networks.
\newblock {\em Scientific Reports}, 15(1):1160, 2025.

\bibitem{gao2024efficient}
Xin Gao, Tianheng Qiu, Xinyu Zhang, Hanlin Bai, Kang Liu, Xuan Huang, Hu~Wei, Guoying Zhang, and Huaping Liu.
\newblock Efficient multi-scale network with learnable discrete wavelet transform for blind motion deblurring.
\newblock In {\em Proceedings of the IEEE/CVF Conference on Computer Vision and Pattern Recognition}, pages 2733--2742, 2024.

\bibitem{tian2023freedyg}
Yuxing Tian, Yiyan Qi, and Fan Guo.
\newblock Freedyg: Frequency enhanced continuous-time dynamic graph model for link prediction.
\newblock In {\em The Twelfth International Conference on Learning Representations}, 2024.

\bibitem{necula2012transient}
V~Necula, S~Klimenko, and G~Mitselmakher.
\newblock Transient analysis with fast wilson-daubechies time-frequency transform.
\newblock In {\em Journal of Physics: Conference Series}, volume 363, page 012032. IOP Publishing, 2012.

\bibitem{DBLP:conf/iclr/TrivediFBZ19}
Rakshit Trivedi, Mehrdad Farajtabar, Prasenjeet Biswal, and Hongyuan Zha.
\newblock Dyrep: Learning representations over dynamic graphs.
\newblock In {\em 7th International Conference on Learning Representations}. OpenReview.net, 2019.

\bibitem{gravina2024long}
Alessio Gravina, Giulio Lovisotto, Claudio Gallicchio, Davide Bacciu, and Claas Grohnfeldt.
\newblock Long range propagation on continuous-time dynamic graphs.
\newblock {\em In Proceedings of the 41st International Conference on Machine Learning}, 2025.

\bibitem{ding2024dygmamba}
Zifeng Ding, Yifeng Li, Yuan He, Antonio Norelli, Jingcheng Wu, Volker Tresp, Michael Bronstein, and Yunpu Ma.
\newblock Dygmamba: Efficiently modeling long-term temporal dependency on continuous-time dynamic graphs with state space models.
\newblock {\em TMLR}, 2025.

\end{thebibliography}
	% Reproducibility Checklist
\appendix
%\newpage
\section{Appendix:Implementation Details}
%\subsection{Details of Implementation}
%All simulations are conducted on a machine with  Intel(R) Xeon(R) Gold 6326 CPU @ 2.90GHz, and NVIDIA RTX A6000. For the construction and operation of the algorithm, it is realized through Pytorch. The source code of our TFWaveFormer is provided in the supplementary materials to ensure reproducibility.

\subsection{Details of Datasets}
We evaluate the proposed method across ten real-world datasets from diverse domains, including Wikipedia, Reddit, MOOC, Social Evolution, LastFM, Enron, UCI, Flights, Contact, and UN Trade, covering a broad spectrum of applications such as online collaboration, social interaction, education, communication, mobility, and international trade~\cite{yu2023towards}. To comprehensively assess the scalability and robustness of our approach, we categorize these datasets into two groups based on their structural complexity and scale characteristics.

Small-scale datasets encompass four networks with relatively compact structures: Enron (184 nodes, 125K edges), UN Trade (255 nodes, 507K edges), Contact (692 nodes, 2.4M edges), and UCI (1,899 nodes, 59K edges). Despite their smaller node counts, these datasets exhibit varying temporal patterns and edge densities, with some featuring high interaction frequencies over extended periods (e.g., UN Trade spans 32 years, while Contact captures intensive proximity interactions within one month).

Large-scale datasets comprise six networks with substantial structural complexity: LastFM (1,980 nodes, 1.3M edges), MOOC (7,144 nodes, 411K edges), Wikipedia (9,227 nodes, 157K edges), Reddit (10,984 nodes, 672K edges), Flights (13,169 nodes, 1.9M edges), and Social Evolution (74 nodes, 2.1M edges). These datasets present diverse challenges in terms of node connectivity, temporal dynamics, and domain-specific characteristics, ranging from social media interactions to transportation networks and educational platforms.

This systematic evaluation across both small-scale and large-scale networks enables us to demonstrate the versatility and effectiveness of our proposed TFWaveFormer method across different graph sizes, temporal patterns, and application domains, providing comprehensive insights into its performance characteristics and scalability properties.

We evaluate our approach on ten carefully selected datasets that span diverse domains and exhibit varying network structural characteristics, ensuring comprehensive assessment of our method's generalizability.

Social Media and Collaboration Networks: \textbf{Wikipedia} represents a bipartite user-article editing network with rich 172-dimensional LIWC linguistic features and temporal edit labels. \textbf{Reddit} captures cross-subreddit user posting dynamics with similar linguistic feature representations, enabling analysis of community interaction patterns.

Educational and Entertainment Platforms: \textbf{MOOC} models student-content engagement in online learning environments through 4-dimensional behavioral features. \textbf{LastFM} documents temporal music listening behaviors, focusing purely on user-artist interaction patterns without explicit edge attributes.

Communication Networks: This category encompasses three distinct communication modalities: \textbf{Enron} traces corporate email exchanges over a three-year period, providing insights into organizational communication dynamics; \textbf{Social Evolution} monitors face-to-face proximity interactions within a university dormitory setting using 2-dimensional spatial features; and \textbf{UCI} captures peer-to-peer messaging patterns among university students.

Infrastructure and Economic Networks: \textbf{Flights} represents a weighted aviation network during the COVID-19 pandemic, capturing disrupted air traffic patterns between airports. \textbf{UN Trade} encompasses agricultural commodity trade relationships among 181 nations spanning three decades, offering insights into long-term economic evolution.

Physical Proximity Networks: \textbf{Contact} provides high-resolution analysis of physical proximity among 700 university students, with fine-grained 5-minute temporal granularity for detailed social interaction modeling.

\begin{table}[h]
	\centering
	\caption{Dataset Statistics.}
	\resizebox{0.46\textwidth}{!}{
		\begin{tabular}{lrrrr}
			\toprule
			\textbf{Dataset} & \textbf{\#Nodes} & \textbf{\#Links} & \textbf{Time Span} & \textbf{Domain} \\
			\hline
			Wikipedia & 9,227 & 157,474 & 1 month & Social \\
			Reddit & 10,984 & 672,447 & 1 month & Social \\
			MOOC & 7,144 & 411,749 & 17 months & Interaction \\
			LastFM & 1,980 & 1,293,103 & 1 month & Interaction \\
			Enron & 184 & 125,235 & 3 years & Social \\
			Social Evo. & 74 & 2,099,519 & 8 months & Proximity \\
			UCI & 1,899 & 59,835 & 196 days & Social \\
			Flights & 13,169 & 1,927,145 & 4 months & Transport \\
			UN Trade & 255 & 507,497 & 32 years & Economics \\
			Contact & 692 & 2,426,279 & 1 month & Proximity \\
			\bottomrule
	\end{tabular}}
	\label{tab:dataset_stats}
\end{table}

\subsection{Details of Experiment Results}
We evaluate on three increasingly difficult negative sampling strategies: historical (hist), and inductive (ind). The detailed results are shown in Appendix Tables 5, 6, 7, 8, our model performs strongly even under the more challenging hist and ind settings, demonstrating robust generalization to unseen links and evolving structures.

\begin{figure}[t]
	\centering
\begin{subfigure}{0.3\textwidth}
		\includegraphics[width=\textwidth]{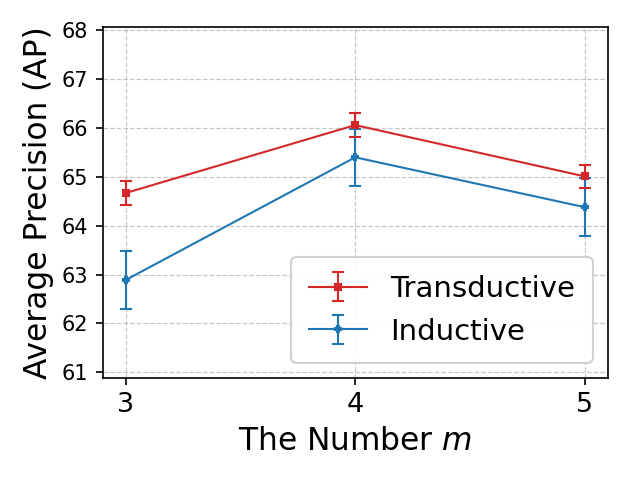}
		\caption{}
	\end{subfigure}
\begin{subfigure}{0.3\textwidth}
		\includegraphics[width=\textwidth]{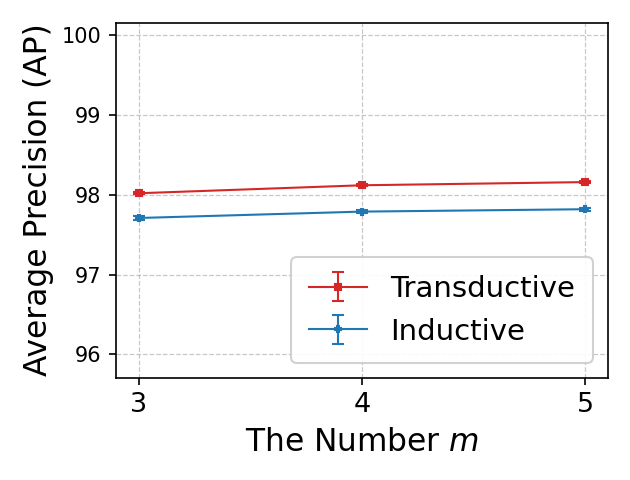}
		\caption{}
	\end{subfigure}
	\caption{Results of hyper-parameters sensitivity study to the number of wavelet convolution kernels $m$ in dynamic link prediction experiments across different datasets. (a) UN Trade, (b) Contact.}
	\label{fig:wavelet_kernel}
\end{figure}
\subsection{Details of Parameter analysis}
In Figure 5, we provide experimental results on the remaining benchmark datasets to complement the main findings presented in the paper. These extended results further validate the robustness and generalizability of our approach across different network structures and scales, reinforcing the conclusions drawn in the main text.
\begin{table*}[h]
	\centering
	\caption{AP for transductive dynamic link prediction with historical, and inductive negative sampling strategies. NSS is the abbreviation of Negative Sampling Strategies.}
	\resizebox{0.8\textwidth}{!}
	{
		\setlength{\tabcolsep}{0.9mm}
		{\renewcommand{\arraystretch}{1.5}
			\begin{tabular}{c|c|ccccccccccc}
				\hline
				NSS                    & Datasets    & JODIE        & DyRep        & TGAT         & TGN          & CAWN         & EdgeBank     & TCL          & GraphMixer &FreeDyG  & DyGFormer & TFWaveFormer    \\ \hline
				\multirow{10}{*}{hist} 
				& Wikipedia   & 83.01 $\pm$ 0.66 & 79.93 $\pm$ 0.56 & 87.38 $\pm$ 0.22 & 86.86 $\pm$ 0.33 & 71.21 $\pm$ 1.67 & 73.35 $\pm$ 0.00 & \underline{89.05 $\pm$ 0.39} & \textbf{90.90 $\pm$ 0.10}& 76.13$\pm$1.65& 82.23 $\pm$ 2.54&86.21$\pm$1.09 \\
				& Reddit      & 80.03 $\pm$ 0.36 & 79.83 $\pm$ 0.31 & 79.55 $\pm$ 0.20 & 81.22 $\pm$ 0.61 & 80.82 $\pm$ 0.45 & 73.59 $\pm$ 0.00 & 77.14 $\pm$ 0.16 & 78.44 $\pm$ 0.18 &\underline{83.08$\pm$0.50}& 81.57 $\pm$ 0.67&\textbf{83.77$\pm$0.43} \\
				& MOOC        & 78.94 $\pm$ 1.25 & 75.60 $\pm$ 1.12 & 82.19 $\pm$ 0.62 & \underline{87.06 $\pm$ 1.93} & 74.05 $\pm$ 0.95 & 60.71 $\pm$ 0.00 & 77.06 $\pm$ 0.41 & 77.77 $\pm$ 0.92 &80.59$\pm$0.29 &85.85 $\pm$ 0.66&\textbf{88.31$\pm$0.37} \\
				& LastFM      & 74.35 $\pm$ 3.81 & 74.92 $\pm$ 2.46 & 71.59 $\pm$ 0.24 & 76.87 $\pm$ 4.64 & 69.86 $\pm$ 0.43 & 73.03 $\pm$ 0.00 & 59.30 $\pm$ 2.31 & 72.47 $\pm$ 0.49 &\underline{83.41$\pm$0.67}& 81.57 $\pm$ 0.48&\textbf{84.80$\pm$0.45} \\
				& Enron       & 69.85 $\pm$ 2.70 & 71.19 $\pm$ 2.76 & 64.07 $\pm$ 1.05 & 73.91 $\pm$ 1.76 & 64.73 $\pm$ 0.36 & 76.53 $\pm$ 0.00 & 70.66 $\pm$ 0.39 & 77.98 $\pm$ 0.92 &\textbf{78.87$\pm$0.82}& 75.63 $\pm$ 0.73&\underline{78.64$\pm$0.35} \\
				& Social Evo. & 87.44 $\pm$ 6.78 & 93.29 $\pm$ 0.43 & 95.01 $\pm$ 0.44 & 94.45 $\pm$ 0.56 & 85.53 $\pm$ 0.38 & 80.57 $\pm$ 0.00 & 94.74 $\pm$ 0.31 & 94.93 $\pm$ 0.31 &\textbf{97.79$\pm$0.23}& \underline{97.38 $\pm$ 0.14}&96.96$\pm$0.21 \\
				& UCI         & 75.24 $\pm$ 5.80 & 55.10 $\pm$ 3.14 & 68.27 $\pm$ 1.37 & 80.43 $\pm$ 2.12 & 65.30 $\pm$ 0.43 & 65.50 $\pm$ 0.00 & 80.25 $\pm$ 2.74 & 84.11 $\pm$ 1.35 &\underline{89.17$\pm$0.31} &82.17 $\pm$ 0.82&\textbf{90.19$\pm$0.51} \\
				& Flights     & 66.48 $\pm$ 2.59 & 67.61 $\pm$ 0.99 & \textbf{72.38 $\pm$ 0.18} & 66.70 $\pm$ 1.64 & 64.72 $\pm$ 0.97 & 70.53 $\pm$ 0.00 & 70.68 $\pm$ 0.24 
				& \underline{71.47 $\pm$ 0.26} &66.03$\pm$0.57 &66.59 $\pm$ 0.49& 65.31$\pm$0.53\\
				& UN Trade    & 61.39 $\pm$ 1.83 & 59.19 $\pm$ 1.07 & 55.74 $\pm$ 0.91 & 58.44 $\pm$ 5.51 & 55.71 $\pm$ 0.38 & \textbf{81.32 $\pm$ 0.00} & 55.90 $\pm$ 1.17 
				& 57.05 $\pm$ 1.22 &50.00$\pm$0.01 &\underline{64.41 $\pm$ 1.40}& 60.16$\pm$1.22\\
				& Contact     & 95.31 $\pm$ 2.13 & 96.39 $\pm$ 0.20 & 96.05 $\pm$ 0.52 & 93.05 $\pm$ 2.35 & 84.16 $\pm$ 0.49 & 88.81 $\pm$ 0.00 & 93.86 $\pm$ 0.21 & 93.36 $\pm$ 0.41 &97.17$\pm$0.04 &\underline{97.57 $\pm$ 0.06}&\textbf{97.85$\pm$0.02} \\ 
				\cline{2-13} 
				& Avg. Rank   & 6.60&6.90&6.20&5.30&9.80&7.80&6.90&5.30&4.50&\underline{3.80}&\textbf{2.90}\\ \hline
				\multirow{10}{*}{ind}  
				& Wikipedia   & 75.65 $\pm$ 0.79 & 70.21 $\pm$ 1.58 & \underline{87.00 $\pm$ 0.16} & 85.62 $\pm$ 0.44 & 74.06 $\pm$ 2.62 & 80.63 $\pm$ 0.00 & 86.76 $\pm$ 0.72 & \textbf{88.59 $\pm$ 0.17} &77.27$\pm$0.19 &78.29 $\pm$ 5.38&73.36$\pm$4.14 \\
				& Reddit      & 86.98 $\pm$ 0.16 & 86.30 $\pm$ 0.26 & 89.59 $\pm$ 0.24 & 88.10 $\pm$ 0.24 & \textbf{91.67 $\pm$ 0.24} & 85.48 $\pm$ 0.00 & 87.45 $\pm$ 0.29 
				& 85.26 $\pm$ 0.11 &87.81$\pm$0.28&\underline{91.11 $\pm$ 0.40}&90.88$\pm$0.24 \\
				& MOOC        & 65.23 $\pm$ 2.19 & 61.66 $\pm$ 0.95 & 75.95 $\pm$ 0.64 & 77.50 $\pm$ 2.91 & 73.51 $\pm$ 0.94 & 49.43 $\pm$ 0.00 & 74.65 $\pm$ 0.54 & 74.27 $\pm$ 0.92 &73.70$\pm$0.48 &\underline{81.24 $\pm$ 0.69}&\textbf{82.46$\pm$0.66} \\
				& LastFM      & 62.67 $\pm$ 4.49 & 64.41 $\pm$ 2.70 & 71.13 $\pm$ 0.17 & 65.95 $\pm$ 5.98 & 67.48 $\pm$ 0.77 & \underline{75.49 $\pm$ 0.00} & 58.21 $\pm$ 0.89 & 68.12 $\pm$ 0.33 &71.57$\pm$0.59 &73.97 $\pm$ 0.50& \textbf{75.52$\pm$0.67}\\
				& Enron       & 68.96 $\pm$ 0.98 & 67.79 $\pm$ 1.53 & 63.94 $\pm$ 1.36 & 70.89 $\pm$ 2.72 & 75.15 $\pm$ 0.58 & 73.89 $\pm$ 0.00 & 71.29 $\pm$ 0.32 & 75.01 $\pm$ 0.79 &\underline{77.81$\pm$0.65} &77.41 $\pm$ 0.89&\textbf{84.23$\pm$0.51} \\
				& Social Evo. & 89.82 $\pm$ 4.11 & 93.28 $\pm$ 0.48 & 94.84 $\pm$ 0.44 & 95.13 $\pm$ 0.56 & 88.32 $\pm$ 0.27 & 83.69 $\pm$ 0.00 & 94.90 $\pm$ 0.36 & 94.72 $\pm$ 0.33 &\underline{97.57$\pm$0.15} &\textbf{97.68 $\pm$ 0.10}&97.19$\pm$ 0.06 \\
				& UCI         & 65.99 $\pm$ 1.40 & 54.79 $\pm$ 1.76 & 68.67 $\pm$ 0.84 & 70.94 $\pm$ 0.71 & 64.61 $\pm$ 0.48 & 57.43 $\pm$ 0.00 & 76.01 $\pm$ 1.11 & 80.10 $\pm$ 0.51 &\underline{80.68$\pm$0.54} &72.25 $\pm$ 1.71&\textbf{86.95$\pm$0.57} \\
				& Flights     & 69.07 $\pm$ 4.02 & 70.57 $\pm$ 1.82 & \underline{75.48 $\pm$ 0.26} & 71.09 $\pm$ 2.72 & 69.18 $\pm$ 1.52 & \textbf{81.08 $\pm$ 0.00} & 74.62 $\pm$ 0.18 & 74.87 $\pm$ 0.21 &65.08$\pm$1.21 &70.92 $\pm$ 1.78& 71.20$\pm$0.98\\
				& UN Trade    & 60.42 $\pm$ 1.48 & 60.19 $\pm$ 1.24 & 60.61 $\pm$ 1.24 & 61.04 $\pm$ 6.01 & \underline{62.54 $\pm$ 0.67} & \textbf{72.97 $\pm$ 0.00} & 61.06 $\pm$ 1.74 & 60.15 $\pm$ 1.29 &50.00$\pm$0.01 &55.79 $\pm$ 1.02& 61.37$\pm$1.17\\
				& Contact     & 93.43 $\pm$ 1.78 & 94.18 $\pm$ 0.10 & 94.35 $\pm$ 0.48 & 90.18 $\pm$ 3.28 & 89.31 $\pm$ 0.27 & 85.20 $\pm$ 0.00 & 91.35 $\pm$ 0.21 & 90.87 $\pm$ 0.35 &\underline{94.91$\pm$0.09} &94.75 $\pm$ 0.28&\textbf{96.10$\pm$0.16} \\
				\cline{2-13} 
				& Avg. Rank  & 8.40&8.90&5.10&5.80&6.90&6.80&5.70&5.90&5.40&\underline{4.20}&\textbf{2.90}   \\ \hline
			\end{tabular}
		}
	}
	\label{tab:average_precision_transductive_dynamic_link_prediction}
\end{table*}
\begin{table*}[!htbp]
	\centering
	\caption{AUC-ROC for transductive dynamic link prediction with historical, and inductive negative sampling strategies.}
	\resizebox{0.8\textwidth}{!}
	{
		\setlength{\tabcolsep}{0.9mm}
		{\renewcommand{\arraystretch}{1.5}
			\begin{tabular}{c|c|ccccccccccc}
				\hline
				NSS                    & Datasets    & JODIE        & DyRep        & TGAT         & TGN          & CAWN         & EdgeBank     & TCL          & GraphMixer   &FreeDyG  & DyGFormer & TFWaveFormer  \\ \hline
				\multirow{10}{*}{hist} 
				& Wikipedia   & 80.77 $\pm$ 0.73 & 77.74 $\pm$ 0.33 & 82.87 $\pm$ 0.22 & 82.74 $\pm$ 0.32 & 67.84 $\pm$ 0.64 & 77.27 $\pm$ 0.00 & \underline{85.76 $\pm$ 0.46} & \textbf{87.68 $\pm$ 0.17} &75.17$\pm$1.72 &78.80 $\pm$ 1.95&83.09$\pm$2.03 \\
				& Reddit      & 80.52 $\pm$ 0.32 & 80.15 $\pm$ 0.18 & 79.33 $\pm$ 0.16 & 81.11 $\pm$ 0.19 & 80.27 $\pm$ 0.30 & 78.58 $\pm$ 0.00 & 76.49 $\pm$ 0.16 & 77.80 $\pm$ 0.12 &\underline{81.90$\pm$0.21}&80.54 $\pm$ 0.29&\textbf{82.01$\pm$0.37} \\
				& MOOC        & 82.75 $\pm$ 0.83 & 81.06 $\pm$ 0.94 & 80.81 $\pm$ 0.67 & \underline{88.00 $\pm$ 1.80} & 71.57 $\pm$ 1.07 & 61.90 $\pm$ 0.00 & 72.09 $\pm$ 0.56 & 76.68 $\pm$ 1.40 &81.99$\pm$0.25 &87.04 $\pm$ 0.35&\textbf{89.14$\pm$0.27} \\
				& LastFM      & 75.22 $\pm$ 2.36 & 74.65 $\pm$ 1.98 & 64.27 $\pm$ 0.26 & 77.97 $\pm$ 3.04 & 67.88 $\pm$ 0.24 & 78.09 $\pm$ 0.00 & 47.24 $\pm$ 3.13 & 64.21 $\pm$ 0.73 &78.54$\pm$0.41 &\underline{78.78 $\pm$ 0.35}&\textbf{80.60$\pm$0.27} \\
				& Enron       & 75.39 $\pm$ 2.37 & 74.69 $\pm$ 3.55 & 61.85 $\pm$ 1.43 & 77.09 $\pm$ 2.22 & 65.10 $\pm$ 0.34 & \underline{79.59 $\pm$ 0.00} & 67.95 $\pm$ 0.88 & 75.27 $\pm$ 1.14 &75.74$\pm$0.72 &76.55 $\pm$ 0.52&\textbf{80.30$\pm$0.76}\\
				& Social Evo. & 90.06 $\pm$ 3.15 & 93.12 $\pm$ 0.34 & 93.08 $\pm$ 0.59 & 94.71 $\pm$ 0.53 & 87.43 $\pm$ 0.15 & 85.81 $\pm$ 0.00 & 93.44 $\pm$ 0.68 & 94.39 $\pm$ 0.31 &\textbf{97.42$\pm$0.02} &\underline{97.28 $\pm$ 0.07}&97.19$\pm$0.15\\
				& UCI         & 78.64 $\pm$ 3.50 & 57.91 $\pm$ 3.12 & 58.89 $\pm$ 1.57 & 77.25 $\pm$ 2.68 & 57.86 $\pm$ 0.15 & 69.56 $\pm$ 0.00 & 72.25 $\pm$ 3.46 & 77.54 $\pm$ 2.02 &\underline{84.96$\pm$0.34} &76.97 $\pm$ 0.24&\textbf{85.18$\pm$0.64} \\
				& Flights     & 68.97 $\pm$ 1.87 & 69.43 $\pm$ 0.90 & \underline{72.20 $\pm$ 0.16} & 68.39 $\pm$ 0.95 & 66.11 $\pm$ 0.71 & \textbf{74.64 $\pm$ 0.00} & 70.57 $\pm$ 0.18 & 70.37 $\pm$ 0.23 &68.57$\pm$0.53 &68.09 $\pm$ 0.43&68.63$\pm$0.63 \\
				& UN Trade    & 68.92 $\pm$ 1.40 & 64.36 $\pm$ 1.40 & 60.37 $\pm$ 0.68 & 63.93 $\pm$ 5.41 & 63.09 $\pm$ 0.74 & \textbf{86.61 $\pm$ 0.00} & 61.43 $\pm$ 1.04 
				& 63.20 $\pm$ 1.54 &50.00$\pm$0.01 &\underline{73.86 $\pm$ 1.13}& 69.69$\pm$1.01\\
				& Contact     & 96.35 $\pm$ 0.92 & 96.00 $\pm$ 0.23 & 95.39 $\pm$ 0.43 & 93.76 $\pm$ 1.29 & 83.06 $\pm$ 0.32 & 92.17 $\pm$ 0.00 & 93.34 $\pm$ 0.19 & 93.14 $\pm$ 0.34 &97.10$\pm$0.05 &\underline{97.17 $\pm$ 0.05}&\textbf{97.43$\pm$0.23} \\ 
				\cline{2-13} 
				& Avg. Rank   & 5.30&6.80&7.40&4.90&9.60&6.60&7.50&6.50&5.00&\underline{4.20}&\textbf{2.20}   \\ \hline
				\multirow{10}{*}{ind}  
				& Wikipedia   & 70.96 $\pm$ 0.78 & 67.36 $\pm$ 0.96 & 81.93 $\pm$ 0.22 & 80.97 $\pm$ 0.31 & 70.95 $\pm$ 0.95 & 81.73 $\pm$ 0.00 & \underline{82.19 $\pm$ 0.48} & \textbf{84.28 $\pm$ 0.30} &73.82$\pm$2.57 &75.09 $\pm$ 3.70&75.92$\pm$2.97 \\
				& Reddit      & 83.51 $\pm$ 0.15 & 82.90 $\pm$ 0.31 & \underline{87.13 $\pm$ 0.20} & 84.56 $\pm$ 0.24 & \textbf{88.04 $\pm$ 0.29} & 85.93 $\pm$ 0.00 & 84.67 $\pm$ 0.29 & 82.21 $\pm$ 0.13 &82.39$\pm$0.24 &86.23 $\pm$ 0.51&85.64$\pm$0.56 \\
				& MOOC        & 66.63 $\pm$ 2.30 & 63.26 $\pm$ 1.01 & 73.18 $\pm$ 0.33 & 77.44 $\pm$ 2.86 & 70.32 $\pm$ 1.43 & 48.18 $\pm$ 0.00 & 70.36 $\pm$ 0.37 & 72.45 $\pm$ 0.72 &75.03$\pm$0.45 &\underline{80.76 $\pm$ 0.76}&\textbf{83.16$\pm$0.72} \\
				& LastFM      & 61.32 $\pm$ 3.49 & 62.15 $\pm$ 2.12 & 63.99 $\pm$ 0.21 & 65.46 $\pm$ 4.27 & 67.92 $\pm$ 0.44 & \textbf{77.37 $\pm$ 0.00} & 46.93 $\pm$ 2.59 
				& 60.22 $\pm$ 0.32 &64.12$\pm$0.57 &\underline{69.25 $\pm$ 0.36}&68.37$\pm$0.61 \\
				& Enron       & 70.92 $\pm$ 1.05 & 68.73 $\pm$ 1.34 & 60.45 $\pm$ 2.12 & 71.34 $\pm$ 2.46 & 75.17 $\pm$ 0.50 & 75.00 $\pm$ 0.00 & 67.64 $\pm$ 0.86 & 71.53 $\pm$ 0.85 &\underline{77.27$\pm$0.61} &74.07 $\pm$ 0.64&\textbf{85.34$\pm$0.23} \\
				& Social Evo. & 90.01 $\pm$ 3.19 & 93.07 $\pm$ 0.38 & 92.94 $\pm$ 0.61 & 95.24 $\pm$ 0.56 & 89.93 $\pm$ 0.15 & 87.88 $\pm$ 0.00 & 93.44 $\pm$ 0.72 & 94.22 $\pm$ 0.32 &\textbf{98.47$\pm$0.02} &\underline{97.51 $\pm$ 0.06}&97.34$\pm$0.02 \\
				& UCI         & 64.14 $\pm$ 1.26 & 54.25 $\pm$ 2.01 & 60.80 $\pm$ 1.01 & 64.11 $\pm$ 1.04 & 58.06 $\pm$ 0.26 & 58.03 $\pm$ 0.00 & 70.05 $\pm$ 1.86 & 74.59 $\pm$ 0.74 &\underline{75.13$\pm$0.67} &65.96 $\pm$ 1.18&\textbf{81.00$\pm$0.52} \\
				& Flights     & 69.99 $\pm$ 3.10 & 71.13 $\pm$ 1.55 & \underline{73.47 $\pm$ 0.18} & 71.63 $\pm$ 1.72 & 69.70 $\pm$ 0.75 & \textbf{81.10 $\pm$ 0.00} & 72.54 $\pm$ 0.19 & 72.21 $\pm$ 0.21 &64.26$\pm$1.10 &69.53 $\pm$ 1.17& 71.11$\pm$0.83\\
				& UN Trade    & 66.82 $\pm$ 1.27 & 65.60 $\pm$ 1.28 & 66.13 $\pm$ 0.78 & 66.37 $\pm$ 5.39 & \underline{71.73 $\pm$ 0.74} & \textbf{74.20 $\pm$ 0.00} & 67.80 $\pm$ 1.21 & 66.53 $\pm$ 1.22 &50.00$\pm$0.01 &62.56 $\pm$ 1.51& 70.48$\pm$1.82\\
				& Contact     & 94.47 $\pm$ 1.08 & 94.23 $\pm$ 0.18 & 94.10 $\pm$ 0.41 & 91.64 $\pm$ 1.72 & 87.68 $\pm$ 0.24 & 85.87 $\pm$ 0.00 & 91.23 $\pm$ 0.19 & 90.96 $\pm$ 0.27 &\underline{95.42$\pm$0.11} &95.01 $\pm$ 0.15&\textbf{96.08$\pm$0.09} \\ \cline{2-13} 
				& Avg. Rank   & 7.50&8.50&6.00&5.70&6.60&5.80&6.10&6.10&5.70&\underline{4.90}&\textbf{3.10}\\ \hline
			\end{tabular}
		}
	}
	\label{tab:auc_roc_transductive_dynamic_link_prediction}
\end{table*}			
\begin{table*}[!htbp]
	\centering
	\caption{AP for inductive dynamic link prediction with historical, and inductive negative sampling strategies.}
	\resizebox{0.8\textwidth}{!}
	{
		\setlength{\tabcolsep}{0.9mm}
		{\renewcommand{\arraystretch}{1.5}
			\begin{tabular}{c|c|cccccccccc}
				\hline
				NSS                    & Datasets    & JODIE        & DyRep        & TGAT         & TGN          & CAWN         & TCL          & GraphMixer &FreeDyG  & DyGFormer & TFWaveFormer   \\ \hline
				\multirow{10}{*}{hist} 
				& Wikipedia   & 68.69 $\pm$ 0.39 & 62.18 $\pm$ 1.27 & \underline{84.17 $\pm$ 0.22} & 81.76 $\pm$ 0.32 & 67.27 $\pm$ 1.63 & 82.20 $\pm$ 2.18 & \textbf{87.60 
					$\pm$ 0.30} &68.57$\pm$5.01 &71.42 $\pm$ 4.43&70.08$\pm$5.12 \\
				& Reddit      & 62.34 $\pm$ 0.54 & 61.60 $\pm$ 0.72 & 63.47 $\pm$ 0.36 & 64.85 $\pm$ 0.85 & 63.67 $\pm$ 0.41 & 60.83 $\pm$ 0.25 & 64.50 $\pm$ 0.26 &64.64$\pm$0.38 &\underline{65.37 $\pm$ 0.60}&\textbf{68.50$\pm$0.68} \\
				& MOOC        & 63.22 $\pm$ 1.55 & 62.93 $\pm$ 1.24 & 76.73 $\pm$ 0.29 & 77.07 $\pm$ 3.41 & 74.68 $\pm$ 0.68 & 74.27 $\pm$ 0.53 & 74.00 $\pm$ 0.97 &71.29$\pm$0.57 &\textbf{80.82 $\pm$ 0.30}&\underline{79.97$\pm$0.50}\\
				& LastFM      & 70.39 $\pm$ 4.31 & 71.45 $\pm$ 1.76 & 76.27 $\pm$ 0.25 & 66.65 $\pm$ 6.11 & 71.33 $\pm$ 0.47 & 65.78 $\pm$ 0.65 & 76.42 $\pm$ 0.22 &\underline{77.87$\pm$0.54} &76.35 $\pm$ 0.52&\textbf{79.52$\pm$0.50}\\
				& Enron       & 65.86 $\pm$ 3.71 & 62.08 $\pm$ 2.27 & 61.40 $\pm$ 1.31 & 62.91 $\pm$ 1.16 & 60.70 $\pm$ 0.36 & 67.11 $\pm$ 0.62 & 72.37 $\pm$ 1.37&\underline{73.01$\pm$0.88} & 67.07 $\pm$ 0.62&\textbf{81.26$\pm$0.37} \\
				& Social Evo. & 88.51 $\pm$ 0.87 & 88.72 $\pm$ 1.10 & 93.97 $\pm$ 0.54 & 90.66 $\pm$ 1.62 & 79.83 $\pm$ 0.38 & 94.10 $\pm$ 0.31 & 94.01 $\pm$ 0.47 &\underline{96.69$\pm$0.14} &\textbf{96.82 $\pm$ 0.16}&95.85$\pm$0.24 \\
				& UCI         & 63.11 $\pm$ 2.27 & 52.47 $\pm$ 2.06 & 70.52 $\pm$ 0.93 & 70.78 $\pm$ 0.78 & 64.54 $\pm$ 0.47 & 76.71 $\pm$ 1.00 & 81.66 $\pm$ 0.49& \underline{82.43$\pm$0.28}& 72.13 $\pm$ 1.87&\textbf{87.57$\pm$0.65} \\
				& Flights     & 61.01 $\pm$ 1.65 & 62.83 $\pm$ 1.31 & \underline{64.72 $\pm$ 0.36} & 59.31 $\pm$ 1.43 & 56.82 $\pm$ 0.57 & 64.50 $\pm$ 0.25 & \textbf{65.28 
					$\pm$ 0.24}&56.47$\pm$0.41 & 57.11 $\pm$ 0.21&56.68$\pm$0.31\\
				& UN Trade    & 55.46 $\pm$ 1.19 & \underline{55.49 $\pm$ 0.84} & 55.28 $\pm$ 0.71 & 52.80 $\pm$ 3.19 & 55.00 $\pm$ 0.38 & \textbf{55.76 $\pm$ 1.03} & 54.94 $\pm$ 0.97 &50.00$\pm$0.01 &53.20 $\pm$ 1.07&53.88$\pm$0.94\\
				& Contact     & 90.42 $\pm$ 2.34 & 89.22 $\pm$ 0.66 & \underline{94.15 $\pm$ 0.45} & 88.13 $\pm$ 1.50 & 74.20 $\pm$ 0.80 & 90.44 $\pm$ 0.17 & 89.91 $\pm$ 0.36 &92.33$\pm$0.67 &93.56 $\pm$ 0.52&\textbf{95.27$\pm$0.43} \\ \cline{2-12} 
				& Avg. Rank   & 7.00&7.50&4.80&6.30&7.80&5.00&4.10&5.20&\underline{4.10}&\textbf{3.20}\\ \hline
				\multirow{10}{*}{ind}  
				& Wikipedia   & 68.70 $\pm$ 0.39 & 62.19 $\pm$ 1.28 & \underline{84.17 $\pm$ 0.22} & 81.77 $\pm$ 0.32 & 67.24 $\pm$ 1.63 & 82.20 $\pm$ 2.18 & \textbf{87.60 
					$\pm$ 0.29}&68.57$\pm$4.27 & 71.42 $\pm$ 4.43&70.07$\pm$2.15 \\
				& Reddit      & 62.32 $\pm$ 0.54 & 61.58 $\pm$ 0.72 & 63.40 $\pm$ 0.36 & 64.84 $\pm$ 0.84 & 63.65 $\pm$ 0.41 & 60.81 $\pm$ 0.26 & 64.49 $\pm$ 0.25 &64.62$\pm$0.51 &\underline{65.35 $\pm$ 0.60}&\textbf{68.54$\pm$0.49} \\
				& MOOC        & 63.22 $\pm$ 1.55 & 62.92 $\pm$ 1.24 & 76.72 $\pm$ 0.30 & 77.07 $\pm$ 3.40 & 74.69 $\pm$ 0.68 & 74.28 $\pm$ 0.53 & 73.99 $\pm$ 0.97 &71.31$\pm$0.36 &\textbf{80.82 $\pm$ 0.30}&\underline{79.99$\pm$0.74}\\
				& LastFM      & 70.39 $\pm$ 4.31 & 71.45 $\pm$ 1.75 & 76.28 $\pm$ 0.25 & 69.46 $\pm$ 4.65 & 71.33 $\pm$ 0.47 & 65.78 $\pm$ 0.65 & 76.42 $\pm$ 0.22 &\underline{77.87$\pm$0.42} &76.35 $\pm$ 0.52&\textbf{79.52$\pm$0.51}\\
				& Enron       & 65.86 $\pm$ 3.71 & 62.08 $\pm$ 2.27 & 61.40 $\pm$ 1.30 & 62.90 $\pm$ 1.16 & 60.72 $\pm$ 0.36 & 67.11 $\pm$ 0.62 & 72.37 $\pm$ 1.38 &\underline{72.85$\pm$0.81}& 67.07 $\pm$ 0.62&\textbf{81.26$\pm$0.32} \\
				& Social Evo. & 88.51 $\pm$ 0.87 & 88.72 $\pm$ 1.10 & 93.97 $\pm$ 0.54 & 90.65 $\pm$ 1.62 & 79.83 $\pm$ 0.39 & 94.10 $\pm$ 0.32 & 94.01 $\pm$ 0.47 &\textbf{96.91$\pm$0.12}&\underline{96.82 $\pm$ 0.17}&95.86$\pm$0.30 \\
				& UCI         & 63.16 $\pm$ 2.27 & 52.47 $\pm$ 2.09 & 70.49 $\pm$ 0.93 & 70.73 $\pm$ 0.79 & 64.54 $\pm$ 0.47 & 76.65 $\pm$ 0.99 & 81.64 $\pm$ 0.49 &\underline{82.44$\pm$0.24} &72.13 $\pm$ 1.86&\textbf{87.59$\pm$0.68} \\
				& Flights     & 61.01 $\pm$ 1.66 & 62.83 $\pm$ 1.31 & \underline{64.72 $\pm$ 0.37} & 59.32 $\pm$ 1.45 & 56.82 $\pm$ 0.56 & 64.50 $\pm$ 0.25 & \textbf{65.29 
					$\pm$ 0.24} &56.49$\pm$0.34 &57.11 $\pm$ 0.20& 56.72$\pm$0.15\\
				& UN Trade    & 55.43 $\pm$ 1.20 & 55.42 $\pm$ 0.87 & \underline{55.58 $\pm$ 0.68} & 52.80 $\pm$ 3.24 & 54.97 $\pm$ 0.38 & \textbf{55.66 $\pm$ 0.98} & 54.88 $\pm$ 1.01 &50.00$\pm$0.01 &52.56 $\pm$ 1.70& 53.90$\pm$0.90\\
				& Contact     & 90.43 $\pm$ 2.33 & 89.22 $\pm$ 0.65 & \underline{94.14 $\pm$ 0.45} & 88.12 $\pm$ 1.50 & 74.19 $\pm$ 0.81 & 90.43 $\pm$ 0.17 & 89.91 $\pm$ 0.36 &92.34$\pm$0.41 &93.55 $\pm$ 0.52&\textbf{94.69$\pm$0.49} \\ \cline{2-12} 
				& Avg. Rank   & 6.90&7.70&4.60&6.20&7.80&5.10&4.10&5.10&\underline{4.30}&\textbf{3.20}\\ \hline
			\end{tabular}
		}
	}
	\label{tab:ap_inductive_dynamic_link_prediction}
\end{table*}
\begin{table*}[!htbp]
	\centering
	\caption{AUC-ROC for inductive dynamic link prediction with historical, and inductive negative sampling strategies.}
	\resizebox{0.8\textwidth}{!}
	{
		\setlength{\tabcolsep}{0.9mm}
		{\renewcommand{\arraystretch}{1.5}
			\begin{tabular}{c|c|cccccccccc}
				\hline
				NSS                    & Datasets    & JODIE        & DyRep        & TGAT         & TGN          & CAWN         & TCL          & GraphMixer &FreeDyG  & DyGFormer & TFWaveFormer   \\ \hline
				\multirow{10}{*}{hist} 
				& Wikipedia   & 61.86 $\pm$ 0.53 & 57.54 $\pm$ 1.09 & 78.38 $\pm$ 0.20 & 75.75 $\pm$ 0.29 & 62.04 $\pm$ 0.65 & \underline{79.79 $\pm$ 0.96} & \textbf{82.87 
					$\pm$ 0.21}&65.400$\pm$4.81 & 68.33 $\pm$ 2.82&68.78$\pm$3.00 \\
				& Reddit      & 61.69 $\pm$ 0.39 & 60.45 $\pm$ 0.37 & 64.43 $\pm$ 0.27 & 64.55 $\pm$ 0.50 & \underline{64.94 $\pm$ 0.21} & 61.43 $\pm$ 0.26 & 64.27 $\pm$ 0.13 &63.60$\pm$0.31 &64.81 $\pm$ 0.25&\textbf{66.93$\pm$0.39} \\
				& MOOC        & 64.48 $\pm$ 1.64 & 64.23 $\pm$ 1.29 & 74.08 $\pm$ 0.27 & 77.69 $\pm$ 3.55 & 71.68 $\pm$ 0.94 & 69.82 $\pm$ 0.32 & 72.53 $\pm$ 0.84 &73.60$\pm$0.87 &\underline{80.77 $\pm$ 0.63}&\textbf{81.95$\pm$0.53} \\
				& LastFM      & 68.44 $\pm$ 3.26 & 68.79 $\pm$ 1.08 & 69.89 $\pm$ 0.28 & 66.99 $\pm$ 5.62 & 67.69 $\pm$ 0.24 & 55.88 $\pm$ 1.85 & 70.07 $\pm$ 0.20 &\underline{71.24$\pm$0.41} &70.73 $\pm$ 0.37&\textbf{72.38$\pm$0.34}\\
				& Enron       & 65.32 $\pm$ 3.57 & 61.50 $\pm$ 2.50 & 57.84 $\pm$ 2.18 & 62.68 $\pm$ 1.09 & 62.25 $\pm$ 0.40 & 64.06 $\pm$ 1.02 & 68.20 $\pm$ 1.62 &\underline{70.09$\pm$0.65} &65.78 $\pm$ 0.42&\textbf{81.59$\pm$0.38} \\
				& Social Evo. & 88.53 $\pm$ 0.55 & 87.93 $\pm$ 1.05 & 91.87 $\pm$ 0.72 & 92.10 $\pm$ 1.22 & 83.54 $\pm$ 0.24 & 93.28 $\pm$ 0.60 & 93.62 $\pm$ 0.35 &\textbf{96.94$\pm$0.17} &\underline{96.91 $\pm$ 0.09}&96.52$\pm$ 0.05 \\
				& UCI         & 60.24 $\pm$ 1.94 & 51.25 $\pm$ 2.37 & 62.32 $\pm$ 1.18 & 62.69 $\pm$ 0.90 & 56.39 $\pm$ 0.10 & 70.46 $\pm$ 1.94 & 75.98 $\pm$ 0.84&\underline{76.57$\pm$0.54} & 65.55 $\pm$ 1.01&\textbf{81.35$\pm$0.36} \\
				& Flights     & 60.72 $\pm$ 1.29 & 61.99 $\pm$ 1.39 & \underline{63.38 $\pm$ 0.26} & 59.66 $\pm$ 1.04 & 56.58 $\pm$ 0.44 & \textbf{63.48 $\pm$ 0.23} & 63.30 $\pm$ 0.19&55.03$\pm$0.41 & 56.05 $\pm$ 0.21&57.21$\pm$0.29\\
				& UN Trade    & 58.73 $\pm$ 1.19 & 57.90 $\pm$ 1.33 & 59.74 $\pm$ 0.59 & 55.61 $\pm$ 3.54 & \underline{60.95 $\pm$ 0.80} & \textbf{61.12 $\pm$ 0.97} & 59.88 $\pm$ 1.17&50.00$\pm$0.01 & 58.46 $\pm$ 1.65&59.63$\pm$1.23\\
				& Contact     & 90.80 $\pm$ 1.18 & 88.88 $\pm$ 0.68 & 93.76 $\pm$ 0.41 & 88.84 $\pm$ 1.39 & 74.79 $\pm$ 0.37 & 90.37 $\pm$ 0.16 & 90.04 $\pm$ 0.29 &92.34$\pm$0.49 &\underline{94.14 $\pm$ 0.26}&\textbf{95.51$\pm$0.24} \\ \cline{2-12} 
				& Avg. Rank   & 7.00&8.40&5.00&6.30&7.20&5.20&4.00&5.00&\underline{4.30}&\textbf{2.60}\\ \hline
				\multirow{10}{*}{ind}  
				& Wikipedia   & 61.87 $\pm$ 0.53 & 57.54 $\pm$ 1.09 & 78.38 $\pm$ 0.20 & 75.76 $\pm$ 0.29 & 62.02 $\pm$ 0.65 & \underline{79.79 $\pm$ 0.96} & \textbf{82.88 
					$\pm$ 0.21}&65.40$\pm$4.11 & 68.33 $\pm$ 2.82&68.77$\pm$2.23 \\
				& Reddit      & 61.69 $\pm$ 0.39 & 60.44 $\pm$ 0.37 & 64.39 $\pm$ 0.27 & 64.55 $\pm$ 0.50 & \underline{64.91 $\pm$ 0.21} & 61.36 $\pm$ 0.26 & 64.27 $\pm$ 0.13 &63.56$\pm$0.35 &64.80 $\pm$ 0.25&\textbf{66.97$\pm$0.26} \\
				& MOOC        & 64.48 $\pm$ 1.64 & 64.22 $\pm$ 1.29 & 74.07 $\pm$ 0.27 & 77.68 $\pm$ 3.55 & 71.69 $\pm$ 0.94 & 69.83 $\pm$ 0.32 & 72.52 $\pm$ 0.84 &73.61$\pm$0.59 &\underline{80.77 $\pm$ 0.63}&\textbf{81.96$\pm$0.49} \\
				& LastFM      & 68.44 $\pm$ 3.26 & 68.79 $\pm$ 1.08 & 69.89 $\pm$ 0.28 & 66.99 $\pm$ 5.61 & 67.68 $\pm$ 0.24 & 55.88 $\pm$ 1.85 & 70.07 $\pm$ 0.20 &\underline{71.24$\pm$0.43} &70.73 $\pm$ 0.37&\textbf{72.38$\pm$0.21}\\
				& Enron       & 65.32 $\pm$ 3.57 & 61.50 $\pm$ 2.50 & 57.83 $\pm$ 2.18 & 62.68 $\pm$ 1.09 & 62.27 $\pm$ 0.40 & 64.05 $\pm$ 1.02 & 68.19 $\pm$ 1.63 &\underline{68.79$\pm$0.91} &65.79 $\pm$ 0.42&\textbf{81.59$\pm$0.10} \\
				& Social Evo. & 88.53 $\pm$ 0.55 & 87.93 $\pm$ 1.05 & 91.88 $\pm$ 0.72 & 92.10 $\pm$ 1.22 & 83.54 $\pm$ 0.24 & 93.28 $\pm$ 0.60 & 93.62 $\pm$ 0.35 &\underline{96.79$\pm$0.17} &\textbf{96.91 $\pm$ 0.09}&96.52$\pm$ 0.05 \\
				& UCI         & 60.27 $\pm$ 1.94 & 51.26 $\pm$ 2.40 & 62.29 $\pm$ 1.17 & 62.66 $\pm$ 0.91 & 56.39 $\pm$ 0.11 & 70.42 $\pm$ 1.93 & 75.97 $\pm$ 0.85 &\underline{76.59$\pm$0.54} &65.58 $\pm$ 1.00&\textbf{81.36$\pm$0.35} \\
				& Flights     & 60.72 $\pm$ 1.29 & 61.99 $\pm$ 1.39 & \underline{63.40 $\pm$ 0.26} & 59.66 $\pm$ 1.05 & 56.58 $\pm$ 0.44 & \textbf{63.49 $\pm$ 0.23} & 63.32 $\pm$ 0.19 &55.03$\pm$0.34 &56.05 $\pm$ 0.22& 57.23$\pm$0.21\\
				& UN Trade    & 58.71 $\pm$ 1.20 & 57.87 $\pm$ 1.36 & 59.98 $\pm$ 0.59 & 55.62 $\pm$ 3.59 & \underline{60.88 $\pm$ 0.79} & \textbf{61.01 $\pm$ 0.93} & 59.71 $\pm$ 1.17 &50.00$\pm$0.01 &57.28 $\pm$ 3.06& 59.66$\pm$1.63\\
				& Contact     & 90.80 $\pm$ 1.18 & 88.87 $\pm$ 0.67 & 93.76 $\pm$ 0.40 & 88.85 $\pm$ 1.39 & 74.79 $\pm$ 0.38 & 90.37 $\pm$ 0.16 & 90.04 $\pm$ 0.29 &93.89$\pm$0.19 &\underline{94.14 $\pm$ 0.26}&\textbf{95.30$\pm$0.32} \\ \cline{2-12} 
				& Avg. Rank   & 7.00&8.30&5.00&6.30&7.20&5.20&4.10&5.00&\underline{4.30}&\textbf{2.60}\\ \hline
			\end{tabular}
		}
	}
	\label{tab:auc_roc_inductive_dynamic_link_prediction}
\end{table*}
\end{document}